\documentclass[letterpaper, 10 pt, conference]{ieeeconf}  

\IEEEoverridecommandlockouts                              

\usepackage{xcolor}

\usepackage{graphicx}

\usepackage{gensymb}
\usepackage[hyphens]{url}
\usepackage{multirow}
\usepackage{multicol}
\usepackage{tabularx}
\usepackage{dsfont}
\usepackage{url}
\usepackage{color}
\usepackage{subcaption}
\usepackage{caption}
\usepackage{booktabs}
\usepackage{dcolumn}
\usepackage{amssymb}
\usepackage{hyperref}
\usepackage{amsmath}
\usepackage{colortbl}
\usepackage{tikz}
\colorlet{lightgray}{gray!20}

\usepackage{physics}
\usepackage{graphicx}
\usepackage{float}
\usepackage{subfloat}
\usepackage{longtable}
\usepackage{threeparttable}
\usepackage{xcolor}
\usepackage{diagbox}
\usepackage{cite}
\usepackage{xspace}

\usepackage{color}
\makeatletter
\DeclareRobustCommand\onedot{\futurelet\@let@token\@onedot}
\def\@onedot{\ifx\@let@token.\else.\null\fi\xspace}

\makeatother
\hypersetup{
    colorlinks=true,
    linkcolor=blue,
    filecolor=magenta,      
    urlcolor=cyan,
    pdftitle={Overleaf Example},
    pdfpagemode=FullScreen,
    }

\usepackage{booktabs}
\usepackage{multirow}

 \renewcommand{\paragraph}[1]{
    \vspace{2mm}
     \noindent\textbf{#1} 
 }

\newcommand{\comambaemoji}{\raisebox{-0.2em}{\includegraphics[height=2.0em]{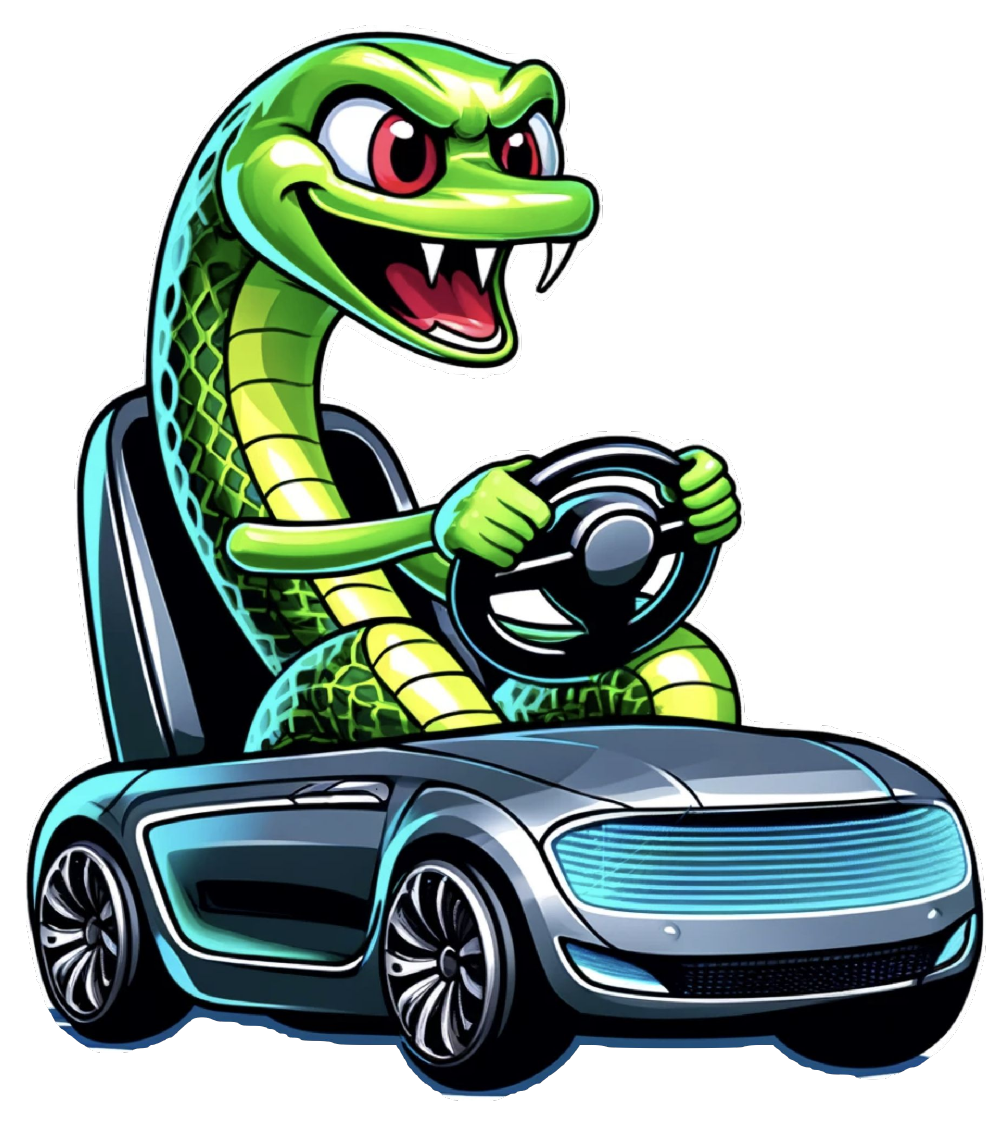}}\xspace}
\definecolor{amber(sae/ece)}{rgb}{1.0, 0.49, 0.0}

\title{ {\LARGE \bf  \comambaemoji \emph{CoMamba}: Real-time Cooperative Perception \\ Unlocked with State Space Models} \\ {\small \href{https://taco-group.github.io/CoMamba/}{https://taco-group.github.io/CoMamba/}}}
\vspace{-3em}

\author{Jinlong Li$^{1}$, Xinyu Liu$^{2}$, Baolu Li$^{2}$,  Runsheng Xu$^{3}$, Jiachen Li$^{4}$, Hongkai Yu$^{2}$,   Zhengzhong Tu$^{1}$
\thanks{$^{1}$Texas A\&M University. $^{2}$Cleveland State University.   $^{3}$University of California, Los Angeles. $^{4}$University of California, Riverside.}
\thanks{*Corresponding Author: tzz@tamu.edu}
}

\begin{document}

 \maketitle
\thispagestyle{empty}
\pagestyle{empty}

\begin{abstract}
Cooperative perception systems play a vital role in enhancing the safety and efficiency of vehicular autonomy.
Although recent studies have highlighted the efficacy of vehicle-to-everything (V2X) communication techniques in autonomous driving, a significant challenge persists: how to efficiently integrate multiple high-bandwidth features across an expanding network of connected agents such as vehicles and infrastructure.
In this paper, we introduce CoMamba, a novel cooperative 3D detection framework designed to leverage state-space models for real-time onboard vehicle perception.
Compared to prior state-of-the-art transformer-based models, CoMamba enjoys being a more scalable 3D model using bidirectional state space models, bypassing the quadratic complexity pain-point of attention mechanisms.
Through extensive experimentation on V2X/V2V datasets, CoMamba achieves superior performance compared to existing methods while maintaining real-time processing capabilities. The proposed framework not only enhances object detection accuracy but also significantly reduces processing time, making it a promising solution for next-generation cooperative perception systems in intelligent transportation networks.
%
\end{abstract}

\vspace{-1.2em}
\section{Introduction}
%
Recently, the new paradigm of cooperative perception~\cite{xu2022opv2v, xu2022cobevt, chen2019f} that engages multiple connected and automated Vehicles (CAVs) has captivated massive research interest.
By leveraging vehicle-to-everything (V2X) or vehicle-to-vehicle (V2V) communication, intelligent actors are now capable of "talking" to their nearby neighbors to share information like pose and sensory data (e.g., point clouds, RGB images, or neural features).
%
%
Although V2X cooperative systems have immense potential to transform the transportation industry, designing efficient fusion strategies to effectively incorporate large, high-dimensional features remains a challenging and unsolved research topic.
Motivated by the phenomenal study on Vision Transformer~\cite{dosovitskiy2020image}, which has demonstrated strong visual learning capabilities on generic vision tasks, prior V2X perception models have been investigating the use of Transformers as the foundational architecture for cooperative perception~\cite{xu2022cobevt,xu2022v2x,li2023s2r,xiang2023hm}.
For example, OPV2V~\cite{xu2022opv2v} implements a single-head self-attention module to fuse features for V2V perception.
V2X-ViT~\cite{xu2022v2x} presents a unified Vision Transformer (ViT) architecture for V2X perception, capable of capturing the heterogeneous nature of V2X systems.
CoBEVT~\cite{xu2022cobevt} proposes a holistic vision Transformer for multi-view cooperative semantic segmentation.
These methods enhance their visual learning capability by leveraging self-attention mechanisms to model long-range spatial interactions.
However, the practical deployment of these methods in large-scale, complex real-world scenarios remains limited due to the slow inference time and worse scalability imbued in attention-based architectures.
\begin{figure*}[!t]
\centering
\includegraphics[width=0.75\textwidth]{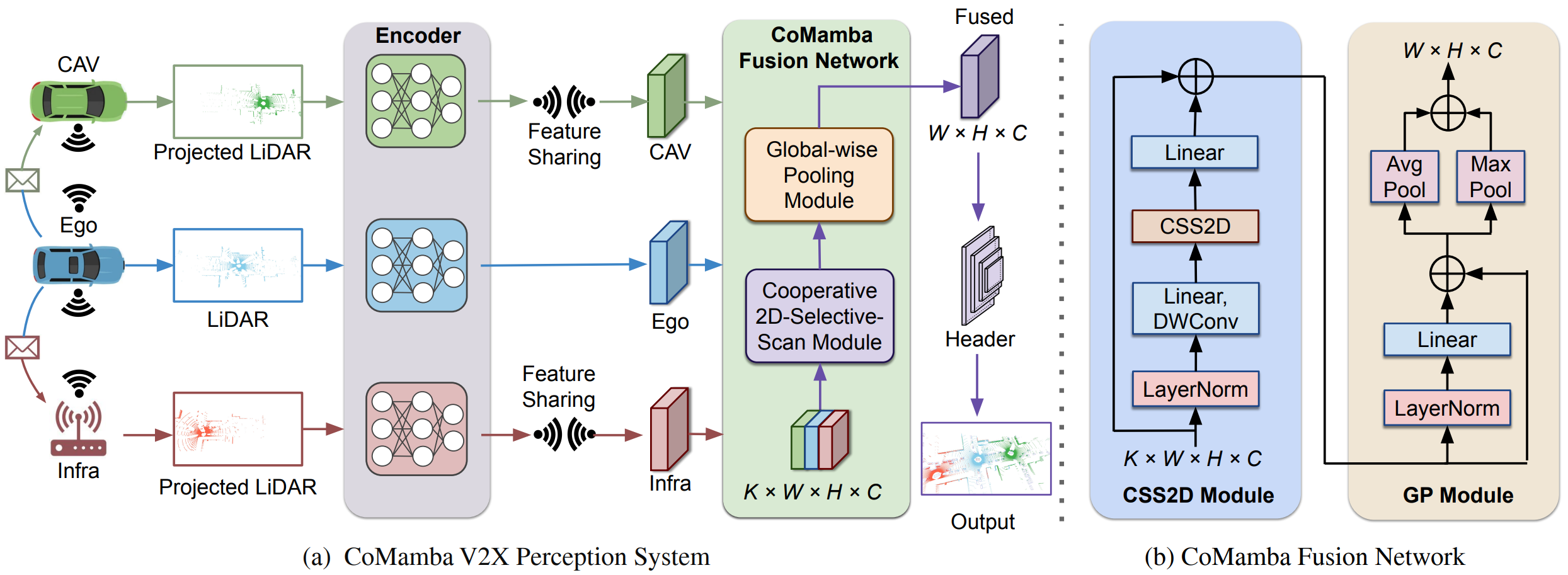}
  \captionsetup{font=small} 
  \caption{\textbf{Overview of our CoMamba V2X-based perception framework.} (a) CoMamba V2X perception system involves V2X metadata sharing, LiDAR visual encoder, feature sharing, and CoMamba fusion network to conduct final prediction. (b) our CoMamba fusion network leverages the Cooperative 2D-Selective-Scan Module to effectively fuse the complex interactions present in high-resource-cost V2X data sequences. The Global-wise Pooling Module efficiently attains global information among the overlapping features of the CAVs.}
  \label{fig:overview}
  \vspace{-1.8em}
\end{figure*}

To overcome these limitations, recent advances in State Space Models (SSMs)~\cite{gu2021efficiently, dao2023hungry, smith2022simplified} offer competitive alternatives to the notoriously compute-intensive Transformers.
A notable model, Mamba~\cite{gu2023mamba}, capably attains long-sequence modeling with a significantly lower \emph{linear complexity} by maintaining a continuous, linear update path through state space.
This efficient design demonstrates impressive performances in long sequence modeling tasks in natural language processing~\cite{zhao2024cobra, qiao2024vl, li2024spikemba}.
Motivated by this success, recent studies have explored its potential for fundamental vision tasks~\cite{ma2024u, huang2024localmamba, liu2024swin, yang2024mambamil}, showcasing impressive performance while considerably fewer computational resources compared to Transformers.
However, while most of these studies predominantly focus on 2D vision tasks such as image recognition, the potential of SSMs to serve as a generic backbone for more challenging vision tasks remains unexplored, particularly in areas involving higher-order visual interaction modeling, such as 3D sequence modeling and space-time interactions.

In this paper, we explore the potential adoption of state-space models for the challenging V2X/V2V cooperative perception task, which involves high-order, multimodal visual information fusion using LiDAR scans.
%
%
We present CoMamba, a generic Mamba-based architecture for efficient V2X cooperative perception that perfectly balances detection performance and computational efficiency.
As illustrated in Fig.~\ref{fig:overview} our CoMamba fusion network comprises two key modules: the Cooperative 2D-Selective-Scan Module and the Global-wise Pooling Module, which we specifically tailored for conducting feature fusion using an intermediate fusion-based cooperative framework.
Together, these modules empower CoMamba to achieve state-of-the-art perception performance on public V2X/V2V perception datasets~\cite{xu2022opv2v,xu2022v2x,xu2023v2v4real}, with significant speed-up as compared to previous transformer-based methods.
Notably, CoMamba unlocks \emph{real-time} cooperative perception with a low latency of 37.1 ms per communication, which translates to 26.9 FPS inference speed with merely a 0.64 GB GPU memory footprint, 19.4\% faster than prior state-of-the-art\footnote{Benchmarked on a single 48GB NVIDIA RTX A6000 card.}.
Our contributions can be summarized as:
\begin{itemize}[leftmargin=*]

    \item We propose CoMamba, the first attempt to explore the potential of linear-complexity Mamba models for V2X cooperative perception. Our CoMamba is a novel V2X perception framework that efficiently models V2X feature interactions using state-space models. Notably, CoMamba scales linearly with the increasing number of connected agents (as explained in Fig.~\ref{fig:cavnum}), whereas previous transformer models all suffer from quadratic complexity with respect to total data dimensionality. 
    
    \item We design two modules inside the CoMamba framework: the Cooperative 2D-Selective-Scan Module for highly efficient 3D spatial interactions and the Global-wise Pooling Module for information aggregation to conduct point cloud-based 3D object detection.
    \item Our comprehensive experimental results on both simulated and real-world datasets have demonstrated that CoMamba exceeds the previous state-of-the-art cooperative detection models while at a significantly lower computational cost. Our ablation studies have shown the efficacy of each component design in contributing to the overall performance.
\end{itemize}

\section{Related Work}\label{Sec:Related_Work}

\noindent \textbf{V2X cooperative perception.}
V2X systems can substantially enhance the perception capabilities of autonomous vehicles by enabling data sharing among CAVs.
This cooperative perception strategy significantly extends the detection range beyond immediate surroundings, thereby improving driving safety in complex scenarios~\cite{lu2023robust,meng2023hydro,liu2020when2com,li2021learning}.
In terms of modeling, V2X-ViT~\cite{xu2022v2x} introduces a unified transformer framework specifically designed to handle the heterogeneous and multi-scale nature of multi-scale V2X systems.
Where2comm~\cite{hu2022where2comm} presents a multi-agent perception framework guided by spatial confidence maps to effectively balance communication bandwidth and perception performance.
CoBEVT~\cite{xu2022cobevt} employs an axial-attention-based multi-agent perception framework that collaboratively generates predictions from sparse locations to capture long-range dependencies.
Additionally, SCOPE~\cite{yang2023spatio} integrates temporal context into a learning-based framework for multi-agent perception to boost the capabilities of the ego agent.
\noindent \textbf{Deployment of V2X perception system.}
Despite the great potential of V2X/V2V systems, deploying these architectures in real-world scenarios requires overcoming numerous fundamental challenges. These include model heterogeneity~\cite{xu2023model,xu2023bridging}, lossy communication~\cite{li2023learning,ren2024interruption}, adversarial vulnerability~\cite{li2024advgps,li2023among}, location errors~\cite{song2023cooperative}, and communication latency~\cite{xu2022v2x, khan2021robust}, to name a few.
Among these, V2X-ViT~\cite{xu2022v2x} introduces a delay-aware positional encoding module to mitigate communication delays and GPS localization errors using a unified Vision Transformer. FDA~\cite{li2024cross} addresses the distribution gap among various private data through a cross-domain learning approach with a feature distribution-aware aggregation framework. S2R-ViT~\cite{li2024s2r} introduces a sim-to-real transfer learning method to reduce the deployment gap affecting V2V perception.\\
\noindent \textbf{State space models}
State space models (SSMs)\cite{gu2021efficiently, dao2023hungry, smith2022simplified}, inspired by linear time-invariant (LTI) systems, emerged as an efficient alternative to transformers for sequence-to-sequence modeling tasks.
One phenomenal model, Mamba\cite{gu2023mamba}, introduces a selection mechanism for dynamically extracting features from sequence data to capture long-range contextual dependencies. Mamba outperforms Transformers on various 1D datasets while requiring significantly fewer computational resources. Motivated by its success in language modeling, state space models have also been extended to various computer vision tasks~\cite{huang2024localmamba, ma2024u, yang2024mambamil, liu2024swin}.
For example, the Visual State-Space Model (Vim)~\cite{zhu2024vision} integrated SSM with bidirectional scanning, enhancing the relational mapping between image patches. VMamba~\cite{liu2024vmamba} further introduces a cross-scan technique, a four-directional modeling approach that uncovers additional spatial connections to fully capture interrelations among image patches.
However, it remains unknown whether SSMs can serve as a new foundation model for more generic vision tasks, such as 3D point cloud understanding, 3D vision, and autonomous driving.

\section{Methodology}\label{Sec:Method}

Current ViT-based V2X perception systems suffer from the quadratic complexity of attention mechanisms as well as large memory footprints, making them impractical for deployment in large-scale, complex real-world scenarios.
Despite some efforts being made to introduce sparse attention for efficiency~\cite{xu2022v2x,xu2022cobevt}, these models fail to scale favorably as the number of agents (or total gathered feature dimensionality in ego vehicle) grows larger. 
We are making the first attempt to explore the potential of linear-complexity Mamba models in the context of V2X cooperative perception to overcome the scalability limitations. 
Inspired by the impressive efficiency and modeling capabilities of SSMs, we build an entirely attention-free architecture, dubbed the CoMamba V2X-based perception framework (illustrated in (a) of Fig.~\ref{fig:overview} ), that is purely based on SSMs. 
Our CoMamba model comprises two major components: the Cooperative 2D-Selective-Scan module and the Global-wise Pooling module.
Thanks to the efficiency-friendly designs in SSMs, our CoMamba model achieves \textbf{real-time inference speed} (26.9 FPS), and scales remarkably better than prior state-of-the-art transformer models.
In this section, we will detail the architectural design of our proposed CoMamba model.

\subsection{Preliminaries}
\noindent \textbf{State space models.}
\label{sec:preliminaries}
State space models (SSMs)~\cite{gu2021efficiently,gu2021combining,smith2022simplified} are continuous sequence-to-sequence modeling systems known for their linear time-invariant (LTI) properties. They map a 1D input sequence $I(x) \in \mathbb{R}$ to a 1D output sequence $O(x) \in \mathbb{R}$ through an intermediate hidden state $h(x) \in \mathbb{R}^{N}$, as illustrated below:
\begin{equation}\label{E1}
\begin{aligned}
h'(x) &= \mathbf{A}h(x) + \mathbf{B}I(x), \;
y(x) &= \mathbf{C}h(x),
\end{aligned}
\end{equation}
where $\mathbf{A} \in \mathbb{R}^{N \times N}$, $\mathbf{B} \in \mathbb{R}^{N \times 1}$, and $\mathbf{C} \in \mathbb{R}^{1 \times N}$ are the evolution and projection parameters, respectively. SSMs effectively capture global system awareness through an implicit mapping to latent states. When $\mathbf{A}$, $\mathbf{B}$, and $\mathbf{C}$ have constant values, Eq.~\ref{E1} defines an LTI system in~\cite{gu2021efficiently}. Otherwise,  Mamba introduces a linear time-varying (LTV) system~\cite{gu2023mamba}. LTI systems inherently lack the ability to perceive content, whereas LTV systems are designed to be input-aware, an important property that attention models also enjoy.  This crucial distinction allows Mamba to surpass the limitations of SSMs, allowing for even stronger modeling capabilities.

To facilitate discretization for deployment in deep learning, a timescale parameter, denoted as $\mathrm{\Delta} \in \mathbb{R}$, is introduced to transform the continuous parameters $\mathbf{A}$ and $\mathbf{B}$ into their discrete counterparts, represented as $\mathbf{\overline{A}}$ and $\mathbf{\overline{B}}$. Using the zero-order hold as the transformation algorithm, the discrete parameters are  formulated as follows:
\begin{equation}\label{E2}
\begin{aligned}
\mathbf{\overline{A}} = \exp(\mathrm{\Delta} \mathbf{A}), \;
\mathbf{\overline{B}} = {(\mathrm{\Delta} \mathbf{A})}^{-1}(\exp(\mathrm{\Delta} \mathbf{A}) - \mathbf{I}) \cdot \mathrm{\Delta} \mathbf{B}.
\end{aligned}
\end{equation}
The discrete form of Eq.~\ref{E1} can then be expressed as:
\begin{equation}\label{E3}
\begin{aligned}
h_x &= \mathbf{\overline{A}}h_{x-1} + \mathbf{\overline{B}}I_{x}, \;
y_x &= \mathbf{C}h_x.
\end{aligned}
\end{equation}

\noindent \textbf{Selective scan mechanism.}
%
Traditional SSMs face limitations due to their LTI properties, resulting in invariant parameters irrespective of variations in the input. To overcome this limitation, the Selective State Space Model (Mamba)~\cite{gu2023mamba} incorporates a selective scan mechanism that integrates three classical techniques: kernel fusion, parallel scan, and recomputation. By employing the selective scan algorithm, Mamba achieves strong modeling capacity while enjoying efficient computational complexity and reduced memory requirements, which contribute to its fast inference.

\begin{figure*}[htb]
\centering
\includegraphics[width=0.8\textwidth]{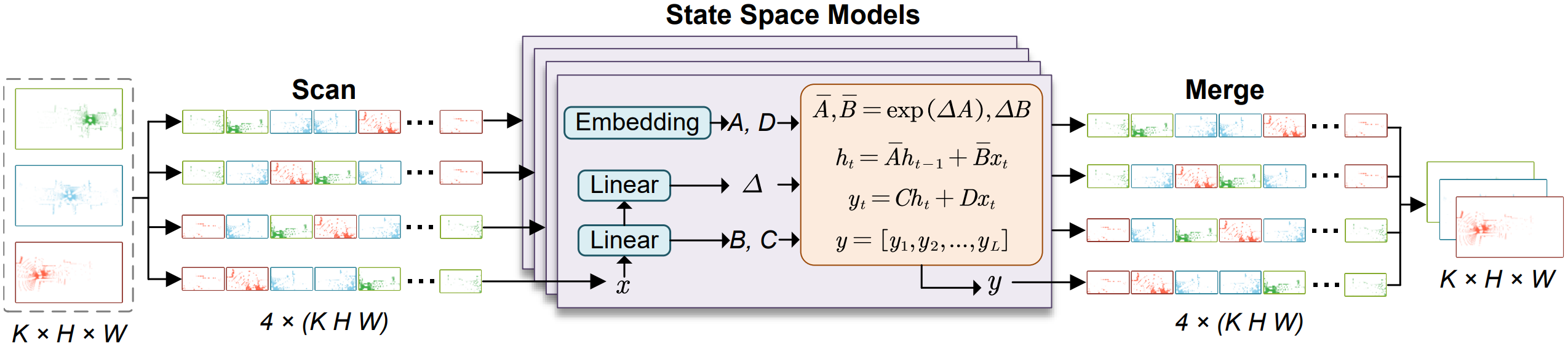}
  \captionsetup{font=small} 
  \caption{\textbf{Illustration of the Cooperative 2D-Selective-Scan (CSS2D) process.}
  The features of $K$ CAVs, represented as $\mathbb{R}^{H \times W}$, are embedded into patches. These patches are then traversed along four different scanning paths, with each 1D sequence ($KHW$) independently processed by distinct Mamba blocks~\cite{gu2023mamba} in parallel. Afterward, the resulting outputs are reshaped and merged to form the 3D feature maps, which maintain the same dimensions as the input features. In this instance, we use \( K = 3 \) as an illustrative example.}
  \label{fig:css2d}
  \vspace{-2.0em}
\end{figure*}

\subsection{CoMamba V2X-based Perception System Design}\label{overview}
The system design of the CoMamba V2X-based perception framework pipeline is illustrated in (a) of Fig.~\ref{fig:overview}.
First, we select an ego vehicle from the CAVs to construct a spatial graph that includes nearby CAVs within the V2X communication sphere. Recognizing the analogous data-sharing capabilities between CAVs and intelligent infrastructures, our methodology equates each infrastructure unit to a CAV.
Then, adjacent CAVs capture and project their raw LiDAR data onto the ego vehicle’s coordinate frame using both their own and the ego vehicle’s GPS positions. The point clouds from the ego vehicle and other CAVs are represented as $\mathbf{P}_{{ego}} \in \mathbb{R}^{4 \times s}$ and $\mathbf{P}_{cav} \in \mathbb{R}^{4 \times s}$, respectively.
In the V2X perception system, each CAV has its own encoder for extracting LiDAR features. After feature extraction, the ego vehicle receives visual features from neighboring CAVs via V2X communication. The intermediate features collected from $N$ surrounding CAVs are denoted as $\mathbf{F}_{cav} \in \mathbb{R}^{N \times H \times W \times C}$, while those of the ego vehicle are denoted as $\mathbf{F}_{ego} \in \mathbb{R}^{1 \times H \times W \times C}$. $\mathbf{F}_{ego}$, along with those $\mathbf{F}_{cav}$ received from other CAVs, are processed by our CoMamba fusion network. The resulting feature map is then passed to a prediction module for 3D bounding-box regression and classification.
Our CoMamba cooperative perception system $\Gamma(\cdot)$ for LiDAR-based 3D object detection can be formulated as follows:
\begin{equation}
\Gamma (\mathbf{P}_{cav}, \mathbf{P}_{ego}) = \Phi(\mathbf{CoMamba}(\mathbf{F}_{cav}, \mathbf{F}_{ego})),
\end{equation}
\begin{equation}
\mathbf{F}_{cav} = \mathbf{E}_{cav}(\mathbf{P}_{cav}), \
\mathbf{F}_{ego} = \mathbf{E}_{ego}(\mathbf{P}_{ego}),
\end{equation}
where $\mathbf{CoMamba}(\cdot)$ is our proposed CoMamba fusion network, which is responsible for efficiently fusing the shared features. $\Phi(\cdot)$ is the prediction header for 3D object detection.
$\mathbf{E}_{ego}$ and $\mathbf{E}_{cav}$ refer to the feature encoders of the ego vehicle and other CAVs, respectively.

\subsection{CoMamba Fusion Network}
\noindent \textbf{Overall architecture.}
The schematic block diagram of the CoMamba fusion network is illustrated in Fig.~\ref{fig:overview}(b).
After encoding by $\mathbf{E}_{ego}$ and $\mathbf{E}_{cav}$, we obtain intermediate neural features $\mathbf{F}_{ego}$ and $\mathbf{F}_{cav}$ from the ego vehicle and other CAVs, respectively. These features are then fed into the  Cooperative 2D-Selective-Scan (CSS2D) module to conduct linear-time 3D information mixing. In the CSS2D module,
We first normalize them by applying Layer Normalization (LN), then followed by a feature extraction using the  $3 \times 3$ depth-wise convolution and Linear layers to obtain their feature maps. 
The processed features are fed into the CSS2D process as shown in Fig.~\ref{fig:css2d} to vision data without compromising its advantages. Then the following features are fed into LN again and the Linear layer with skip-connections, and MaxPool and AvgPool operation modules, which form our Global-wise Pooling Module (GPM) to obtain the final fused feature $\mathbf{F}_{fused} \in \mathbb{R}^{H \times W \times C}$.

\noindent \textbf{Cooperative 2D-Selective-Scan (CSS2D).}
We utilize the four-directional sequence modeling approach proposed in~\cite{liu2024vmamba} to improve global spatial awareness of high-order spatial features.
Specifically, the input feature maps $\mathbf{F}_{ego}$ and $\mathbf{F}_{cav}$ are first flattened into dimensions of $\mathbb{R}^{KHW \times C}$, where $K= N + 1$. This process ensures that all CAVs' neural features within the V2X communication range are flattened into 1D sequence sets $S_{K}$. These 1D sequences are then individually processed through Mamba blocks~\cite{gu2023mamba} for feature extraction to obtain the enhanced 1D sequences $\overline{S}_{K}$. Then we unflatten the outputs  $\overline{S}_{K}$ and combine them to obtain the interactive feature maps  $ \widehat{\mathbf{F}}_{(ego, cav)}  \in \mathbb{R}^{K \times H \times W \times C}$.
The overview of CSS2D process $\mathbf{CSS2D}(\cdot)$ is formulated as
\begin{align}
    \mathbf{CSS2D}(\mathbf{F}_{cav},\mathbf{F}_{cav}) = \mathrm{Merge}(\mathbf{SSM}(\mathrm{Scan}(\mathbf{F}_{ego},\mathbf{F}_{cav})).
    \label{eq:css2d}
\end{align}
where $\mathbf{SSM}(\cdot)$ is the selective scan mechanism~\cite{gu2023mamba}. $\mathrm{Scan} (\cdot)$, and $\mathrm{Merge} (\cdot)$ represent the operation of flattening, and unflattening, respectively.

\noindent \textbf{Global-wise Pooling Module (GPM).}
After being processed by the CSS2D module, the enhanced features are denoted as $\widehat{\mathbf{F}}_{(ego, cav)}$.
To attain global-aware properties among all these CAVs' overlapping features, we utilize the spatial features generated by max pooling and average pooling, which is shown in Fig.~\ref{fig:overview}(b). The  $\widehat{\mathbf{F}}_{(ego, cav)} \in \mathbb{R}^{K \times H \times W \times C}$ features are first fed into the Layer Norm and Linear layer (LLs), then reduced to $ \widehat{\mathbf{F}}_{(ego, cav)}^{max}  \in \mathbb{R}^{1 \times H \times W \times C}$ and $ \widehat{\mathbf{F}}_{(ego, cav)}^{avg}  \in \mathbb{R}^{1 \times H \times W \times C}$ by calculating max pooling and average pooling along the first channel axis.
These two feature maps are combined to get the final fused feature $ \widehat{\mathbf{F}}_{fused}  \in \mathbb{R}^{1 \times H \times W \times C}$  which contains two kinds of global spatial information from the original intermediate feature maps.
This process can be formulated as
\begin{align}
    \mathbf{GPM}(\widehat{\mathbf{F}}_{(ego, cav)}) = \mathrm{P_{max}}(\mathrm{LLs}(\widehat{\mathbf{F}}_{(ego, cav)})) + \\ \mathrm{P_{ave}}(\mathrm{LLs}(\widehat{\mathbf{F}}_{(ego, cav)})),
    \label{eq:GPM}
\end{align}
where $\mathbf{GPM}(\cdot)$ denotes our proposed Global-wise Pooling Module.
$\mathrm{P_{max}} (\cdot)$, and $\mathrm{P_{ave}} (\cdot)$ represent the operation of max pooling and average pooling along the first channel axis, respectively.
For 3D object detection, we use the smooth L1 loss for bounding box regression and focal loss~\cite{lin2017focal} for classification, which constructs our final loss for training.

\begin{figure*}[htb]
\centering
\includegraphics[width=1.8\columnwidth]{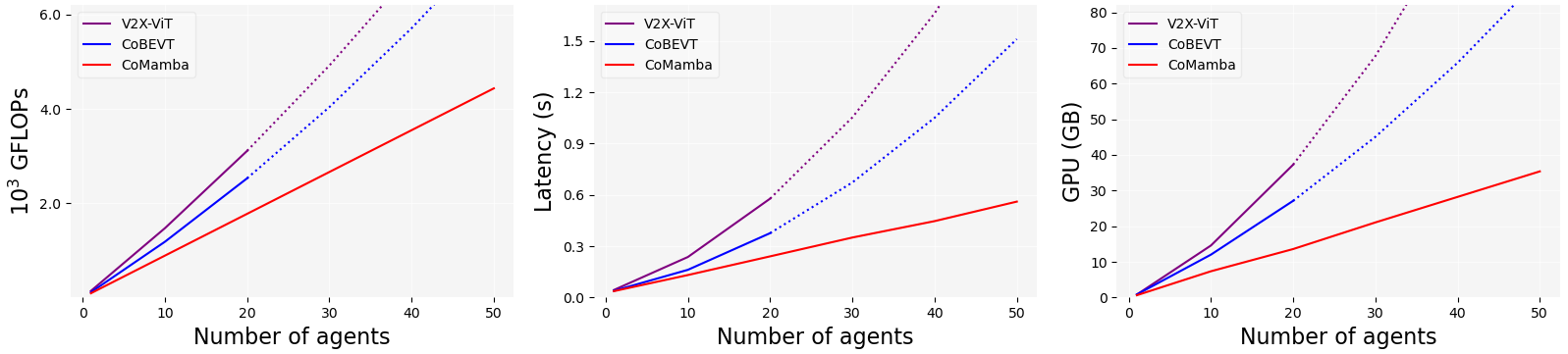}%
\captionsetup{font=small} 
\caption{ \textbf{Scalability comparisons:} CoMamba, V2X-ViT~\cite{xu2022v2x} and CoBEVT~\cite{xu2022cobevt} in GFLOPs, latency, and GPU memory footprint with respect to the number of agents (total feature dimensions). The dotted lines indicate the estimated values when tested GPUs are out-of-memory.
}
\label{fig:cavnum}
\vspace{-2em}
\end{figure*}

\noindent \textbf{Complexity analysis.}
Current V2X  methods are predominantly based on Transformer architectures~\cite{xu2022v2x,xu2022cobevt}.
They have made considerable efforts to optimize spatial computational efficiency but have largely overlooked the potential increased number of connected agents.
With the future proliferation of intelligent agents and V2X perception systems, the scale of agents required for cooperative perception in V2X systems will inevitably grow exponentially.
However, prior cooperative transformers will struggle to handle more CAVs due to self-attention models' quadratic complexity and memory footprint.
We would like to highlight that our proposed CoMamba is truly scalable in terms of the entire spatial dimension, including both 2D feature dimensions and the number of agents.
Fig.~\ref{fig:cavnum} demonstrates the FLOPs, latency, and memory footprint comparisons of our CoMamba against prior state-of-the-art transformer models,
V2X-ViT~\cite{xu2022v2x} and CoBEVT~\cite{xu2022cobevt}. We may see that both Transformer models suffer from quadratic complexity in both metrics, while CoMamba enjoys being linear. When the number of agents exceeds 20, the memory capacity of a single 48GB GPU device (NVIDIA RTX A6000 card) cannot suffice to run the other two models anymore.
In contrast, CoMamba leverages the advantages of SSMs to attain linear costs in GFLOPs, latency, and GPU memory relative to the number of agents, while maintaining excellent performance (Sec.~\ref{ssec:quantitative-evaluation}). 

\section{Experiment}

\subsection{Datasets and experimental setup}
\noindent \textbf{Dataset.}
We conducted extensive experiments on three multi-agent datasets: OPV2V~\cite{xu2022opv2v}, V2XSet~\cite{xu2022v2x}, and V2V4Real~\cite{xu2023v2v4real}. OPV2V~\cite{xu2022opv2v} and V2XSet~\cite{xu2022v2x} are simulated datasets generated using the CARLA simulator and the OpenCDA co-simulation framework~\cite{xu2021opencda}. The OPV2V dataset is organized into 6,764 frames for training, 1,981 frames for validation, and 2,719 frames for testing. Of these, 2,170 frames from CARLA Towns and 594 frames from Culver City are used as two distinct OPV2V testing sets.
\begin{table}[htb]
\captionsetup{font=small} 
\caption{\textbf{LiDAR-based 3D detection performance comparison.} We show Average Precision (AP) at IoU=0.5 and 0.7 on four V2X testing sets from OPV2V, V2X-Set, and V2V4Real datasets.}
\vspace{0.6em}
\resizebox{1.28\columnwidth}{!}{%
\begin{tabular}{@{}c|cc|cc|cc|cc@{}}
\toprule
           & \multicolumn{2}{c|}{OPV2V Default~\cite{xu2022opv2v}} & \multicolumn{2}{c|}{OPV2V Culver City~\cite{xu2022opv2v}} & \multicolumn{2}{c|}{V2XSet~\cite{xu2022v2x}} & \multicolumn{2}{c}{V2V4Real~\cite{xu2023v2v4real}} \\ \midrule
Method     & AP@0.5             & AP@0.7            & AP@0.5             & AP@0.7            & AP@0.5        & AP@0.7       & AP@0.5        & AP@0.7       \\ \midrule
No Fusion  & 67.9               & 60.2              & 55.7               & 47.1              & 60.6          & 40.2         & 39.8          & 22.0         \\
F-Cooper~\cite{chen2019f}   & 86.3               & 75.9              & 81.5               & 71.9              & 84.0          & 68.0         & 53.6          & 26.7         \\
AttFuse~\cite{xu2022opv2v}    & 85.1               & 73.5              & 83.8               & 70.0              & 80.7          & 66.4         & 57.7          & 27.5         \\
V2VAM~\cite{li2023learning}      & 85.7               & 74.3              & 84.1               & 70.9              & 81.3          & 66.1         & 56.8          & 28.1         \\
V2VNet~\cite{wang2020v2vnet}     & 88.1               & 82.2              & 86.8               & 73.4              & 84.5          & 67.7         & 56.4          & 28.5         \\
Where2Comm~\cite{hu2022where2comm} & 89.7               & 80.6              & 84.5               & 65.8              & 85.5          & 72.1         & 58.2          & 28.3         \\
V2X-ViT~\cite{xu2022v2x}    & 87.3               & 72.6              & 87.1               & 72.0              & 88.2          & 71.2         & 55.9          & 29.3         \\
CoBEVT~\cite{xu2022cobevt}     & 90.8               & 82.1              & 86.6               & 74.8              & 84.1          & 71.5         & 58.6          & 29.7         \\
CoMamba (ours)       & \textbf{91.9}               & \textbf{83.3}              & \textbf{87.4}               & \textbf{75.2}              & \textbf{88.3}          & \textbf{72.9}         & \textbf{63.9}          & \textbf{35.5}         \\ \bottomrule
\end{tabular}}
\label{tab:det_result}
\vspace{-2em}
\end{table}
V2XSet is structured into training, validation, and testing segments, with 6,694, 1,920, and 2,833 frames respectively.
V2V4Real~\cite{xu2023v2v4real} is an extensive real-world  V2V perception dataset, collected by two CAVs in Columbus, OH, USA. It contains 20,000 LiDAR frames covering intersections, highway ramps, and urban roads. It is split into 14,210/2,000/3,986 frames for training/validation/testing, respectively.

\noindent \textbf{Compared methods.}
Here, seven state-of-the-art V2X fusion methods are evaluated, all of which employ \textit{Intermediate Fusion} as the primary strategy:  AttFuse~\cite{xu2022opv2v}, V2VNet~\cite{wang2020v2vnet}, 
F-Cooper~\cite{chen2019f}, V2X-ViT~\cite{xu2022v2x}, CoBEVT~\cite{xu2022cobevt}, Where2Comm~\cite{hu2022where2comm}, and V2VAM~\cite{li2023learning}.
We train all these methods on three cooperative perception training sets (\textit{i.e.} OPV2V, V2XSet, and V2V4Real) for a fair comparison. Then, these methods are evaluated using their testing sets to assess their performance.

\noindent \textbf{Evaluation metrics.} 
The final 3D vehicle detection accuracy is selected as our performance evaluation. Following~\cite{xu2022opv2v,xu2022v2x}, we set the evaluation range as $x\in[-140, 140]$ meters,  $y\in[-40, 40]$ meters, where all the CAVs are included in this spatial range in the experiment. We measure the accuracy with Average Precisions (AP) at Intersection-over-Union (IoU) thresholds of $0.5$ and $0.7$. 

\noindent \textbf{Experiment settings.}
To ensure a fair comparison, all methods employ PointPillar~\cite{lang2019pointpillars} as the point cloud encoder. We use the Adam optimizer~\cite{loshchilov2017decoupled} with an initial learning rate of $10^{-3}$, which is gradually decayed every 10 epochs by a factor of 0.1. Following the setup in~\cite{xu2022v2x}, all models are trained on two NVIDIA RTX A6000 GPU cards.
\begin{table}[htb]
\centering
\captionsetup{font=small}
\caption{\textbf{Camera-only 3D detection performance comparison.} We show Average Precision (AP) at IoU=0.5 and 0.7 on the OPV2V and V2XSet datasets.}
\begin{flushright}
\resizebox{0.74\columnwidth}{!}{%
\begin{tabular}{@{}c|cc|cc@{}}
\toprule
\multicolumn{1}{l|}{} & \multicolumn{2}{c|}{OPV2V Default~\cite{xu2022opv2v}} & \multicolumn{2}{c}{V2XSet~\cite{xu2022v2x}} \\ \midrule
Method                & AP@0.5           & AP@0.7          & AP@0.5       & AP@0.7      \\ \midrule
No Fusion             & 45.94            & 25.56           & 30.37        & 13.79       \\
Late Fusion           & 77.62            & 51.92           & 51.41        & 25.59       \\
V2VNet                & 79.06            & 57.59           & 59.54        & 39.00       \\
Where2Comm            & 77.14            & 58.60           & 61.69        & 43.96       \\
V2X-ViT               & 78.41            & 58.38           & 59.14        & 41.23       \\
CoBEVT                & 80.26            & 59.34           & 58.84        & 40.81       \\
CoAlign               & 80.21            & 60.46           & 64.79        & 39.64       \\
CoMamba(ours)         & \textbf{83.12}            & \textbf{63.23}           & \textbf{69.16}        & \textbf{46.58}        \\ \bottomrule
\end{tabular}
}
\end{flushright}
\label{tab:vis_result}
\end{table}
We also conducted extensive experiments on the camera-only cooperative perception task. We utilize the single-scale, history-free BEVFormer as the 3D object detector for individual agents. We employ EfficientNet as the image backbone and use a finer grid resolution of 0.4 meters to preserve detailed spatial information.
\begin{figure*}[htb]
\centering
\subfloat[\footnotesize CoBEVT in OPV2V]{%
  \includegraphics[width=0.4\columnwidth]{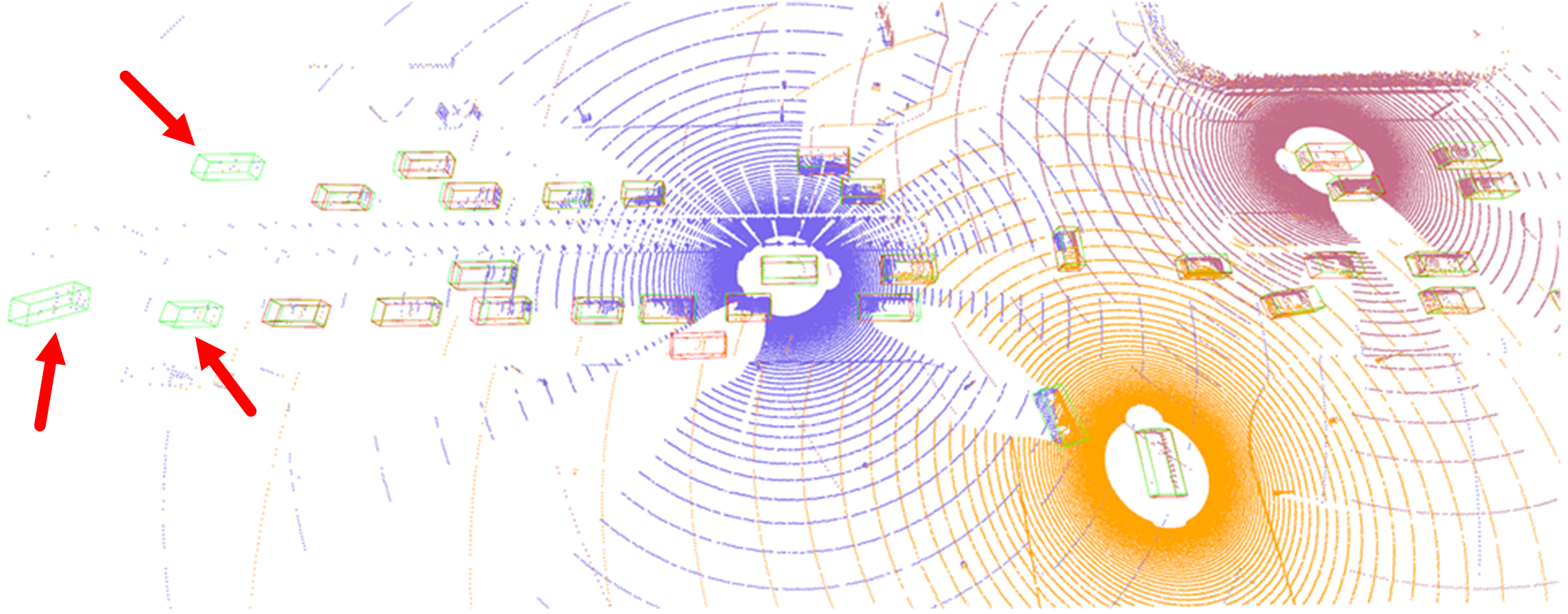}%
}
 \hfil
\subfloat[\footnotesize V2X-ViT in OPV2V]{%
  \includegraphics[width=0.4\columnwidth]{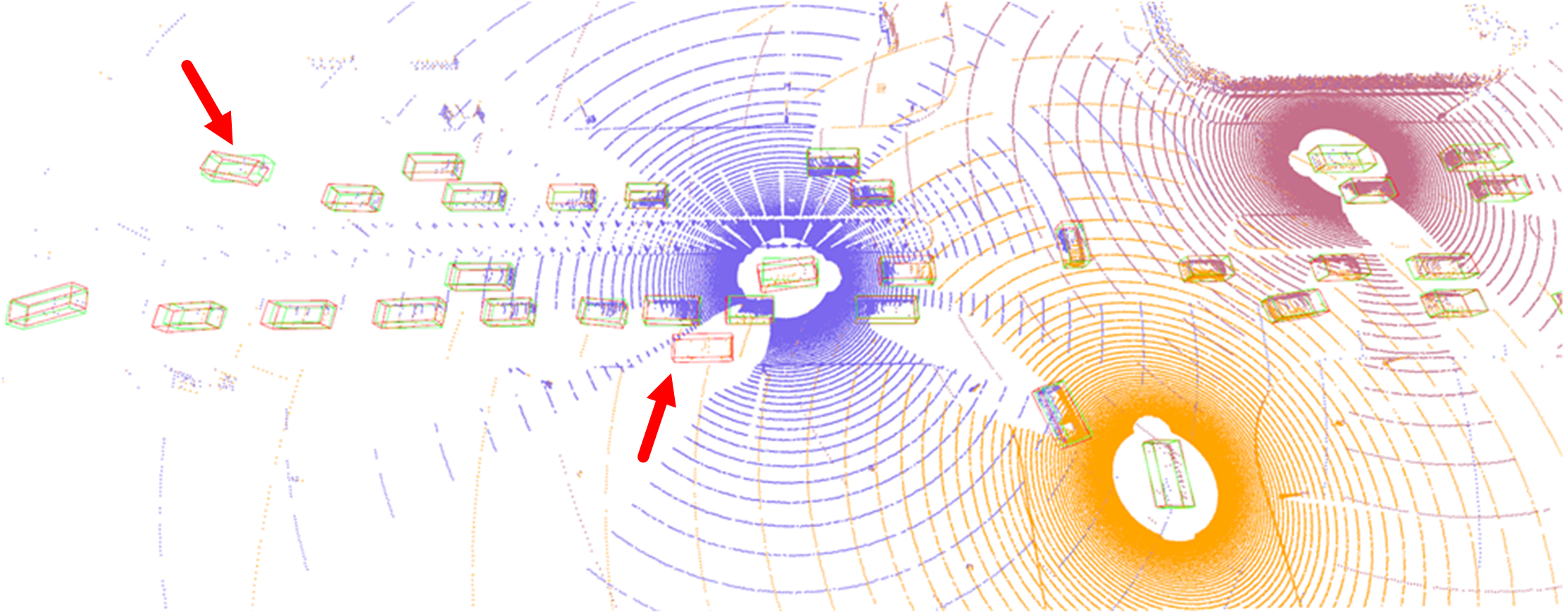}%
}
 \hfil
\subfloat[\footnotesize CoMamba in OPV2V]{%
  \includegraphics[width=0.4\columnwidth]{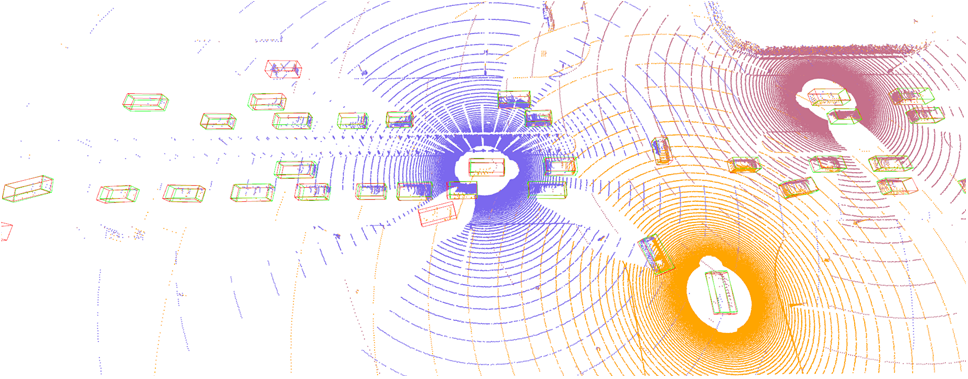}%
}
\\ 
\subfloat[\footnotesize CoBEVT in V2XSet]{%
  \includegraphics[width=0.4\columnwidth]{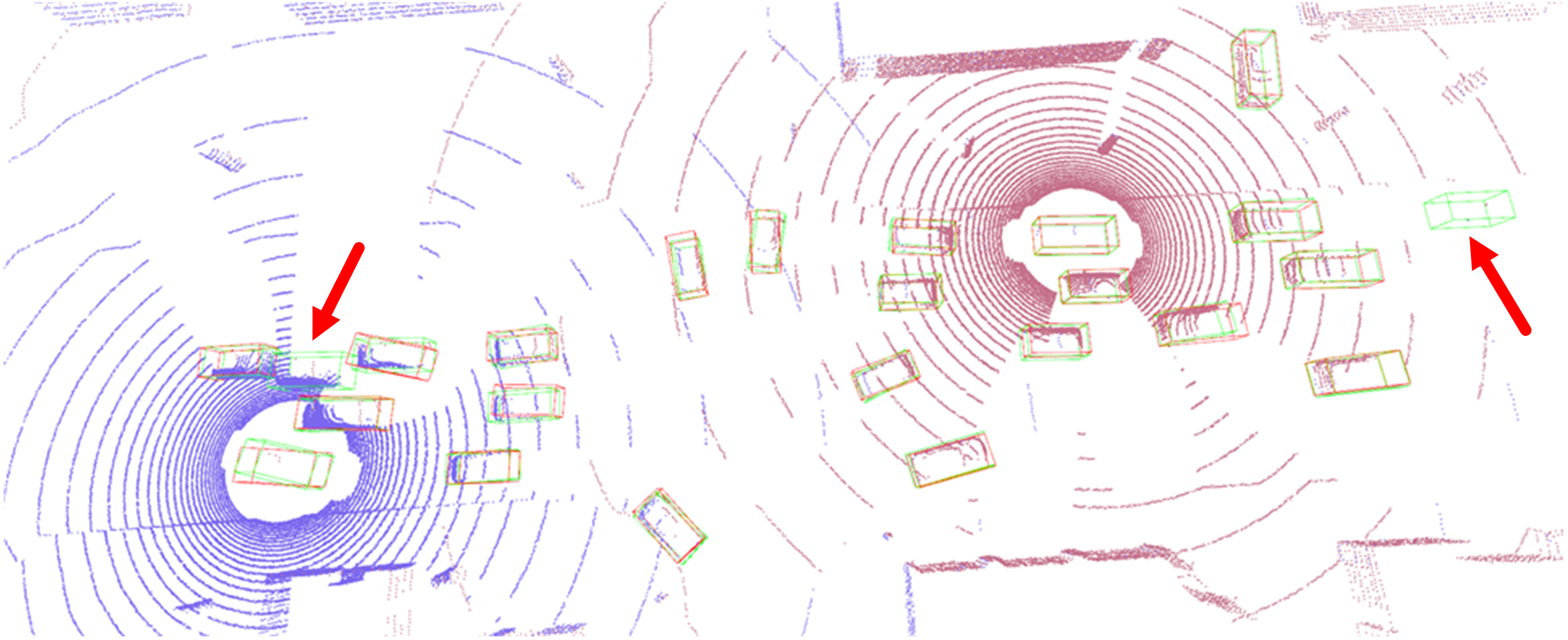}%
}
 \hfil
\subfloat[\footnotesize V2X-ViT in V2XSet]{%
  \includegraphics[width=0.4\columnwidth]{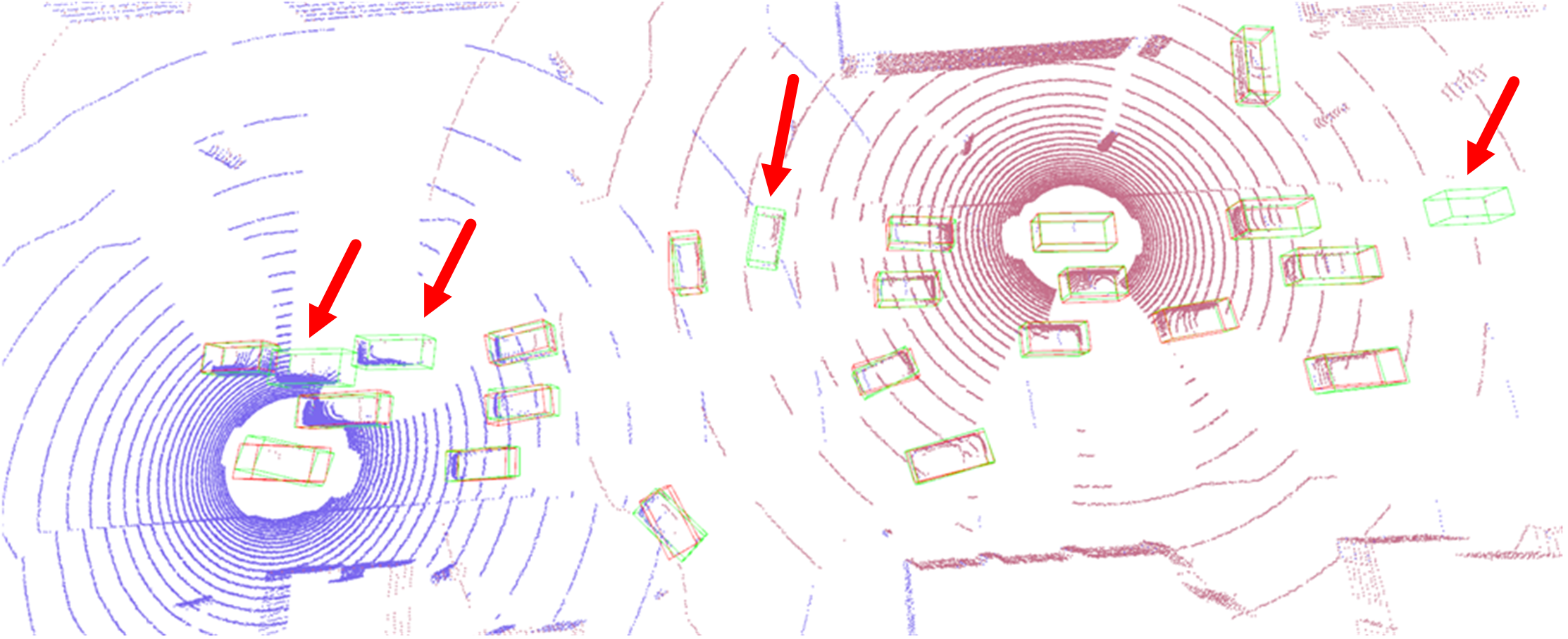}%
}
 \hfil
\subfloat[\footnotesize CoMamba in V2XSet]{%
  \includegraphics[width=0.4\columnwidth]{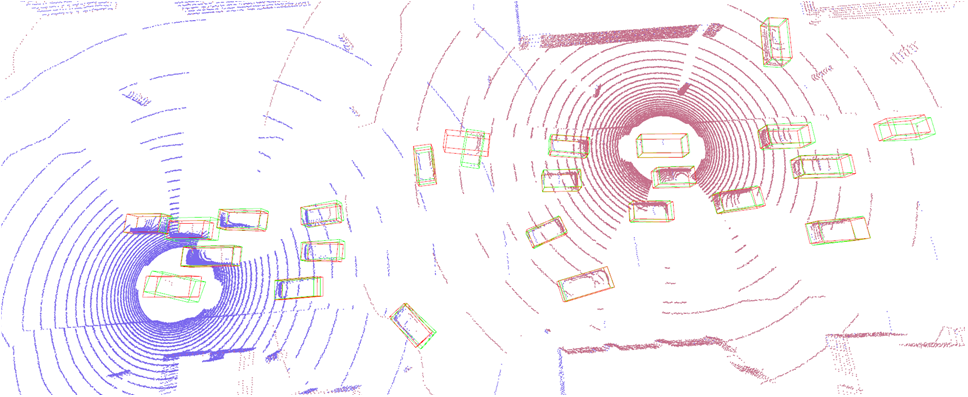}%
}
\\ 
\subfloat[\footnotesize CoBEVT in V2V4Real]{%
  \includegraphics[width=0.4\columnwidth]{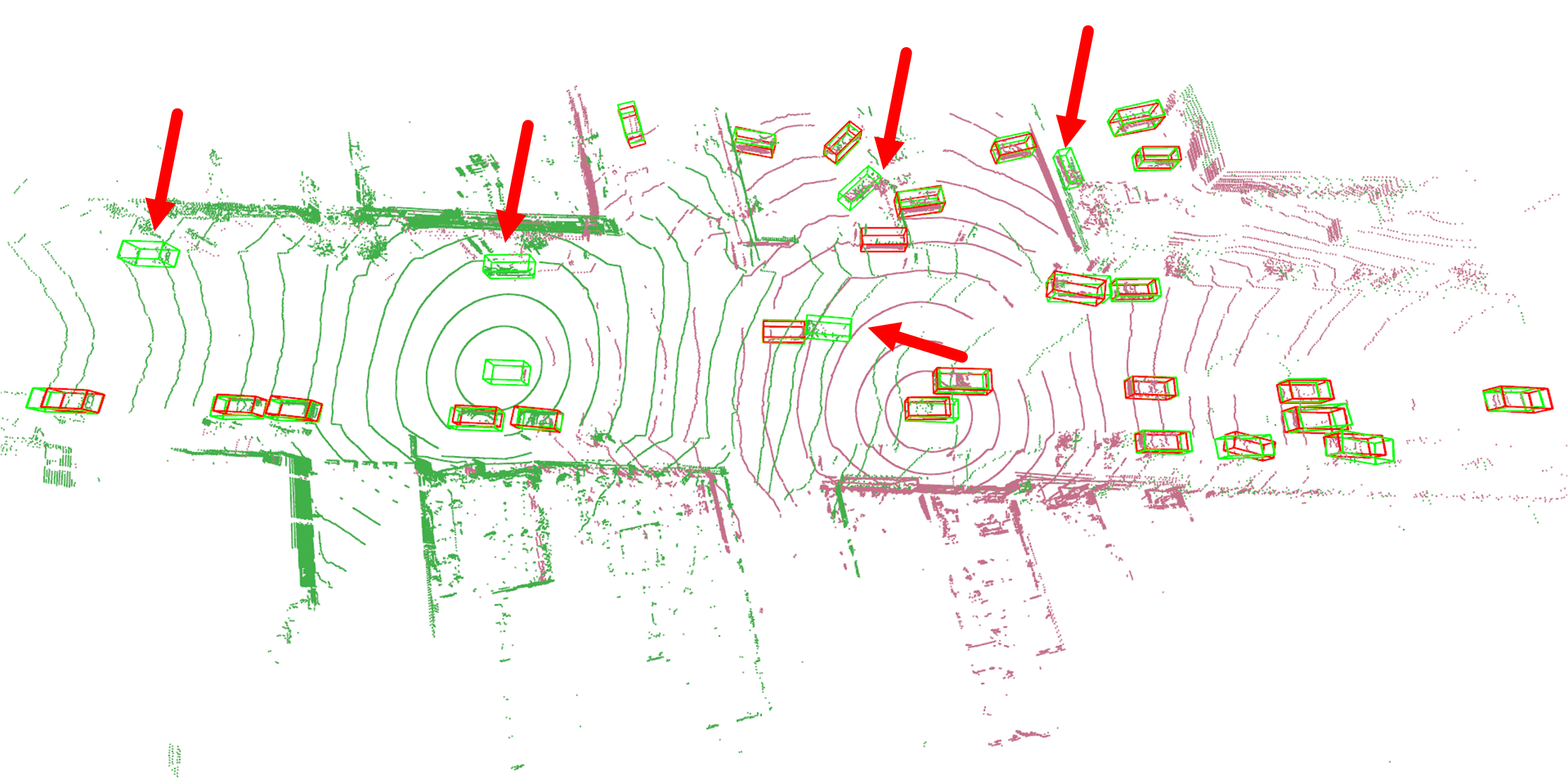}%
}
 \hfil
\subfloat[\footnotesize V2X-ViT in V2V4Real]{%
  \includegraphics[width=0.4\columnwidth]{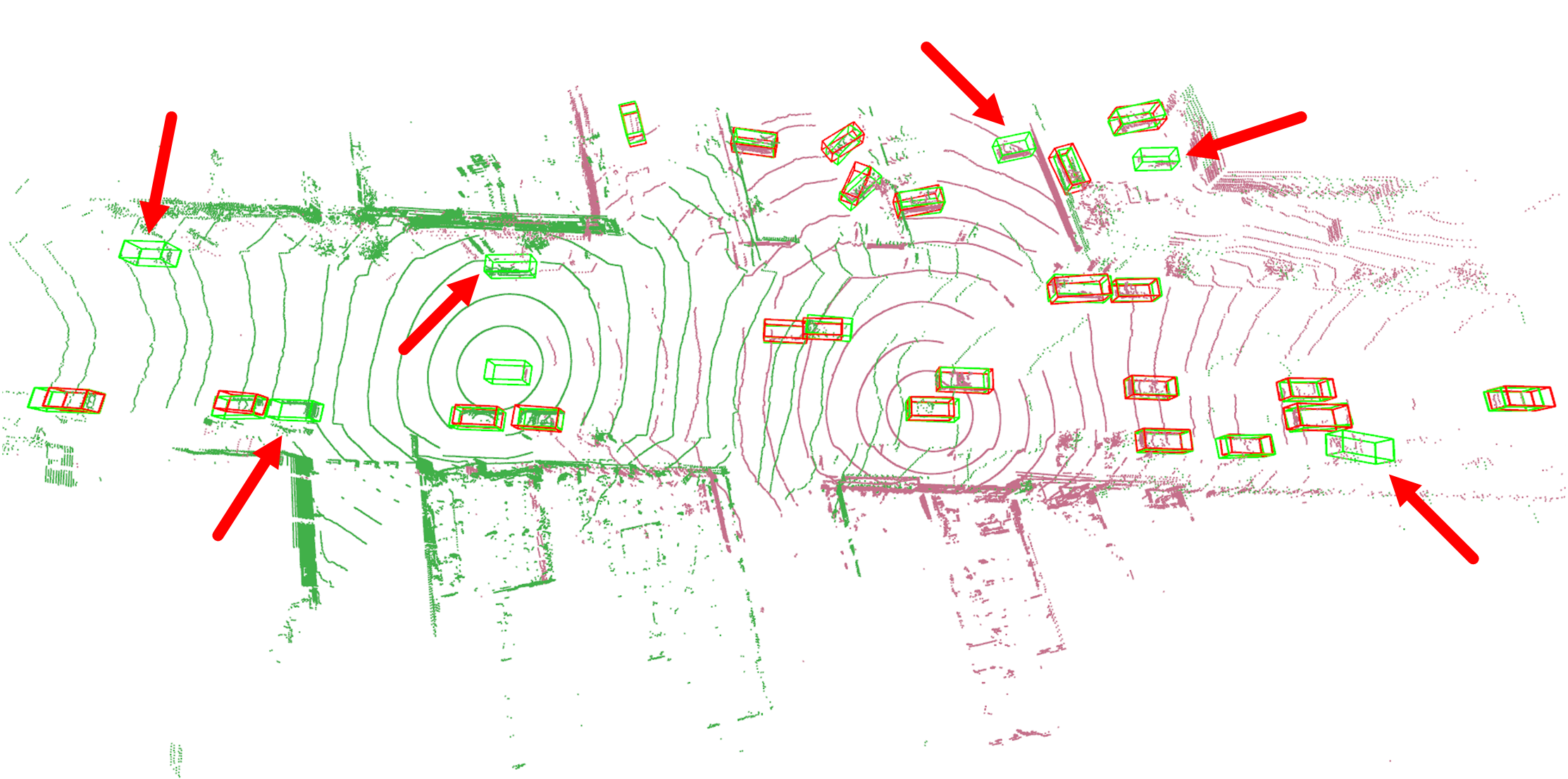}%
}
 \hfil
\subfloat[\footnotesize CoMamba in V2V4Real]{%
  \includegraphics[width=0.4\columnwidth]{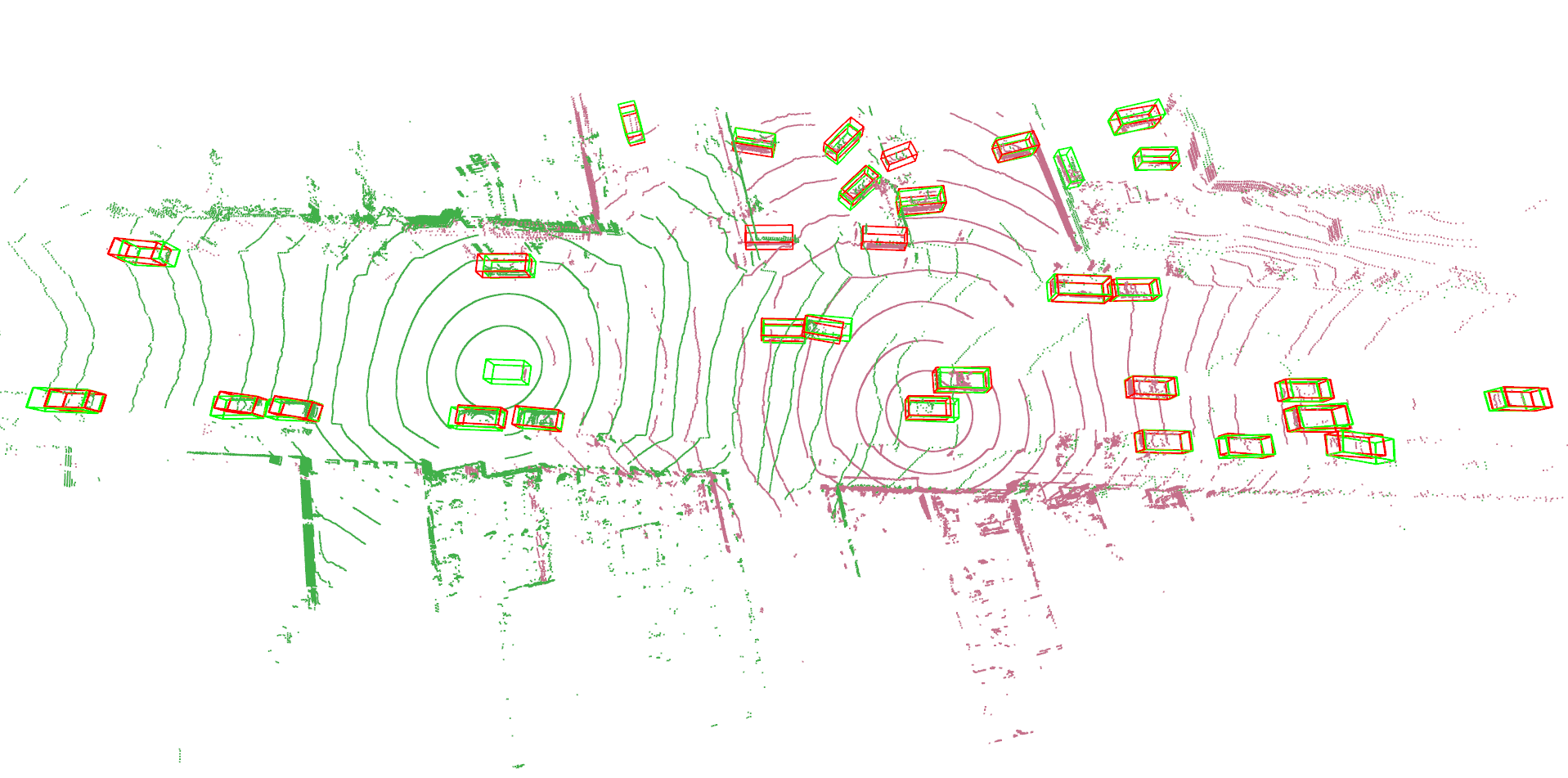}%
}
\captionsetup{font=small} 
\caption{\textbf{Visualizations of 3D object detection results.} \textcolor{green}{Green} and \textcolor{red}{red} 3D bounding boxes represent the ground truth and prediction, respectively. Some false detection examples are highlighted using the red arrow.}
\label{fig:v2x_vis}
\vspace{-1.7em}
\end{figure*}

\begin{figure}[!t]
\centering
\subfloat[\scriptsize point cloud in s-1]{%
  \includegraphics[width=0.45\columnwidth]{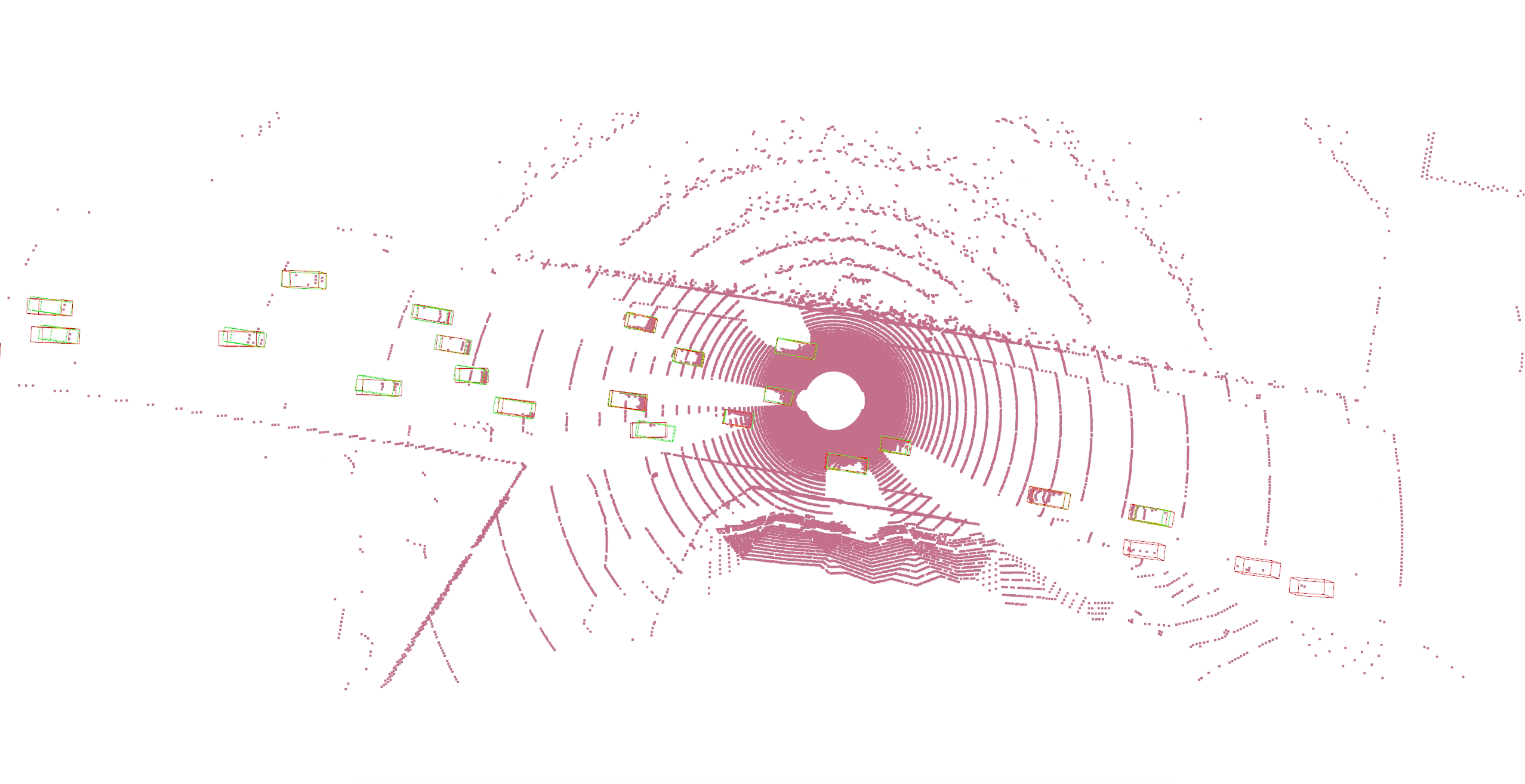}%
}
 \hfil
\subfloat[\scriptsize point cloud in s-2]{%
  \includegraphics[width=0.45\columnwidth]{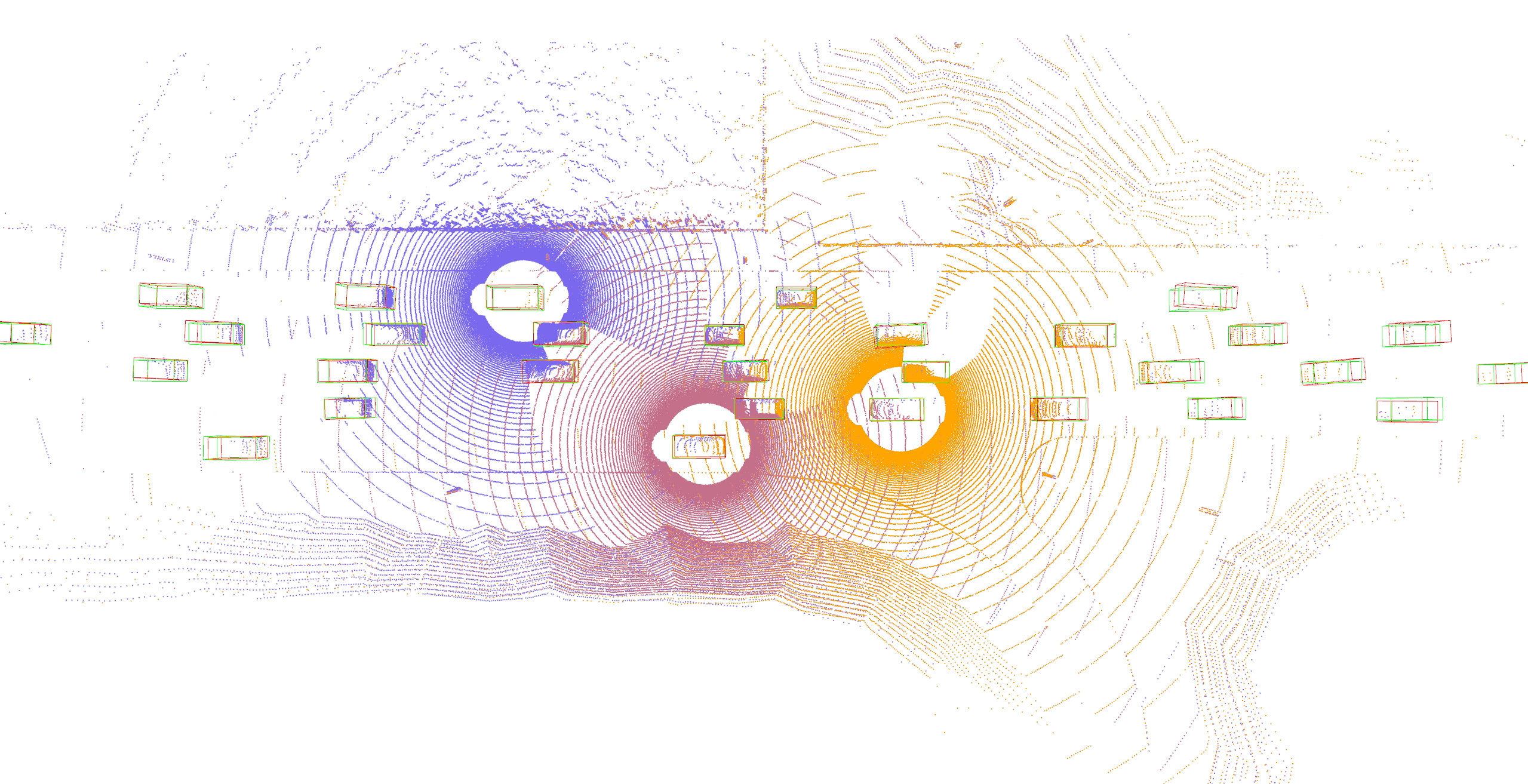}%
}
\\
\subfloat[\scriptsize CoBEVT in s-1]{%
  \includegraphics[width=0.45\columnwidth]{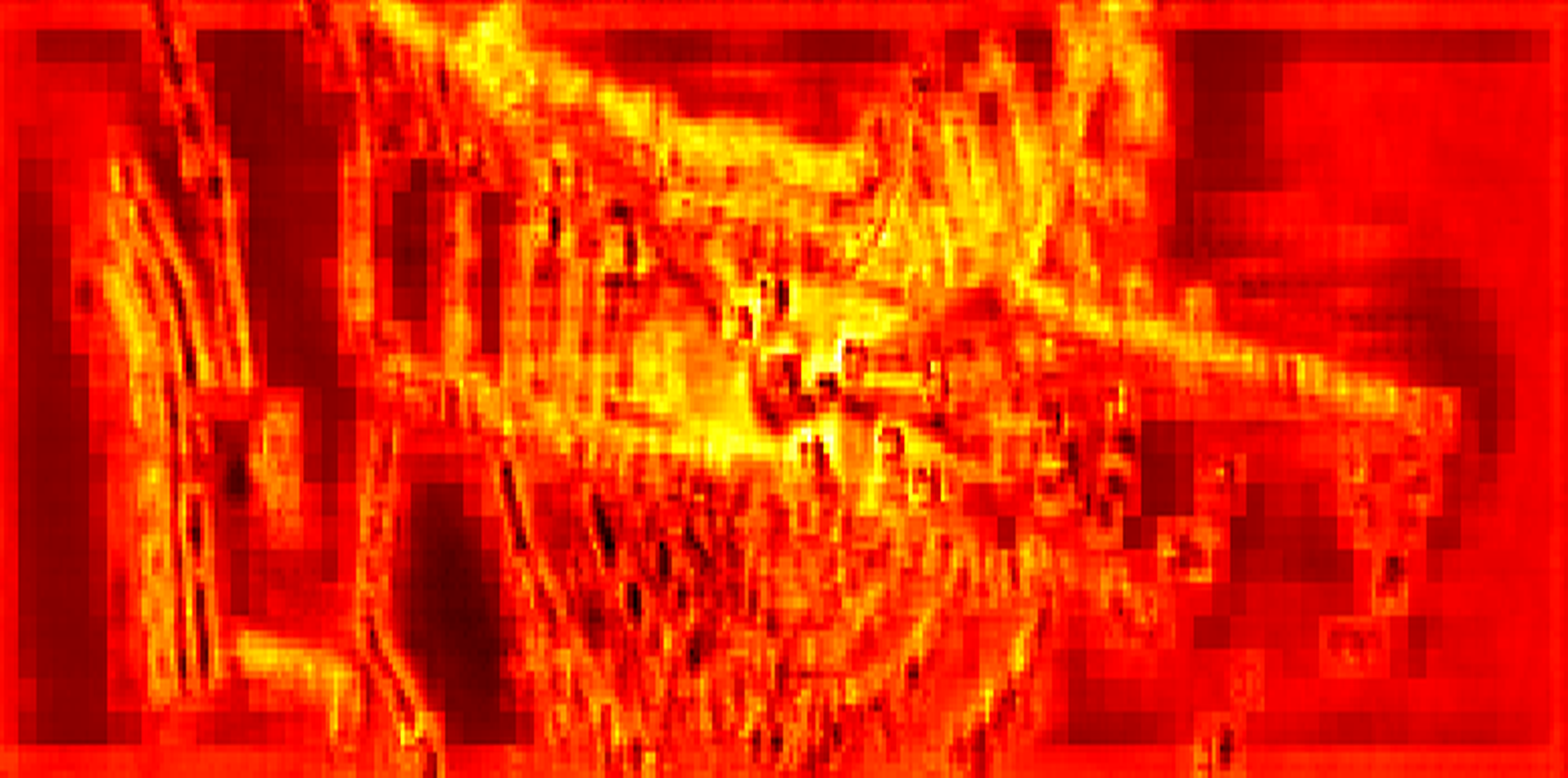}%
}
 \hfil
\subfloat[\scriptsize CoBEVT in s-2]{%
  \includegraphics[width=0.45\columnwidth]{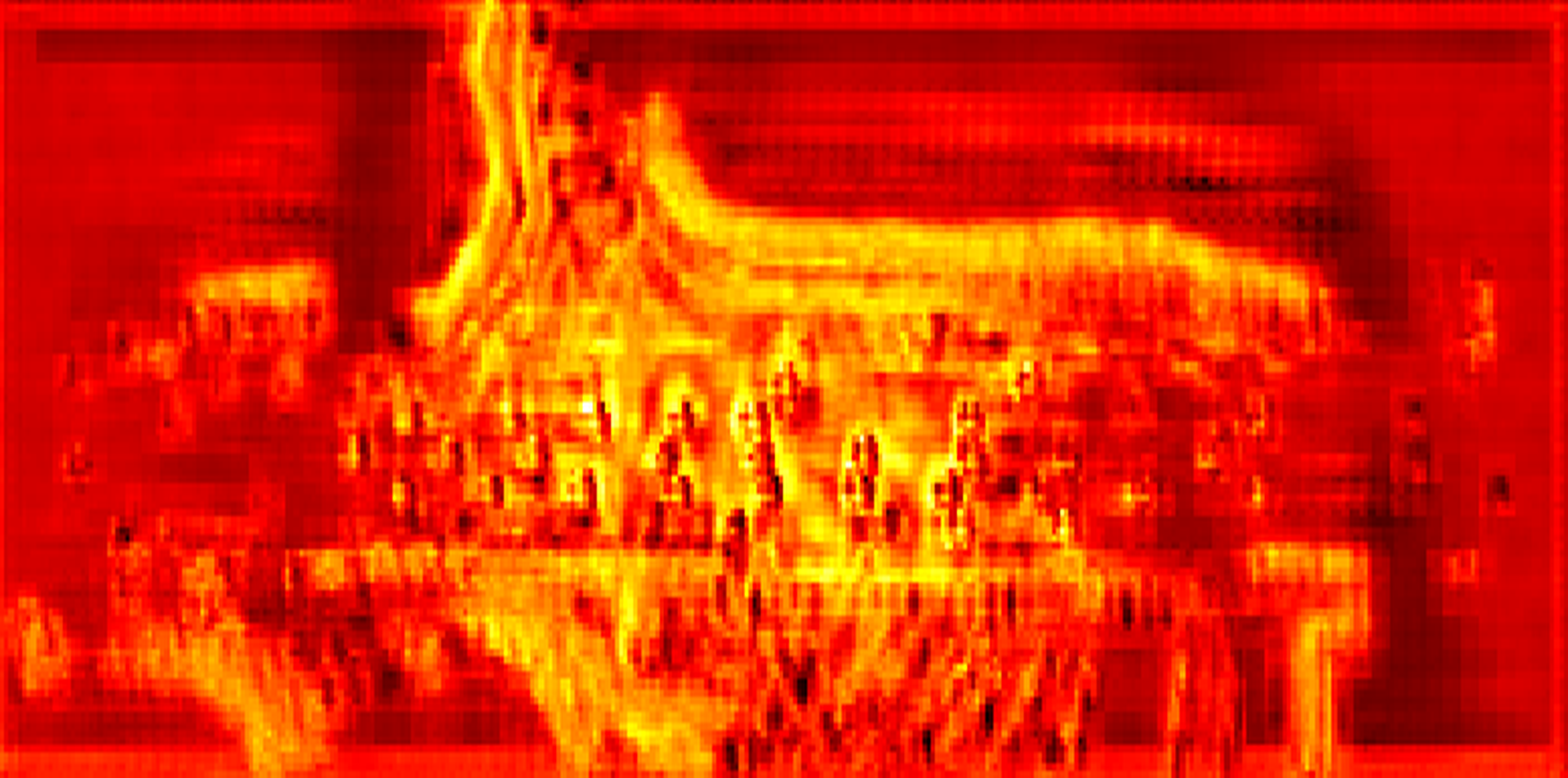}%
}
\\
\subfloat[\scriptsize V2X-ViT in s-1]{%
  \includegraphics[width=0.45\columnwidth]{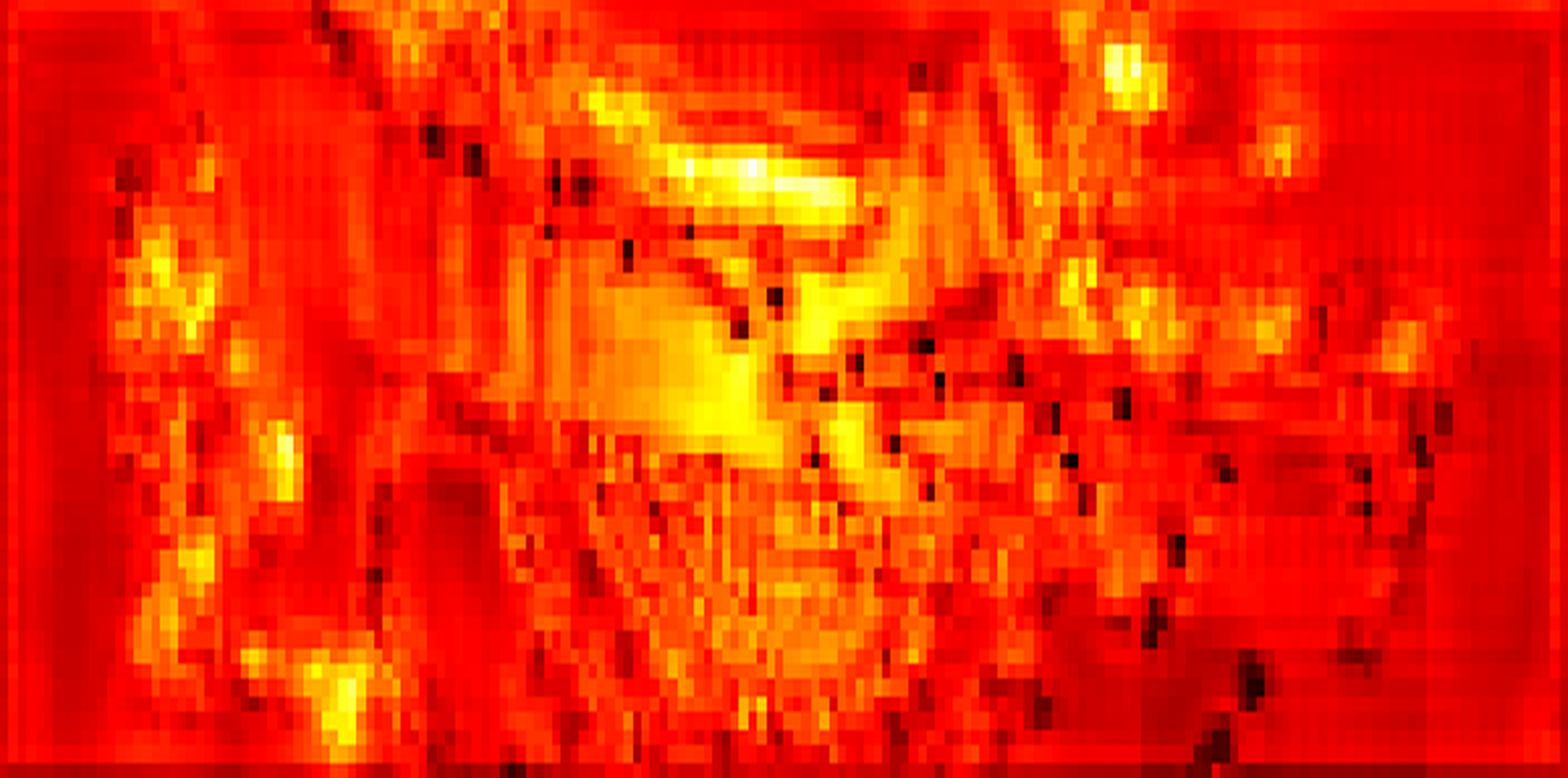}%
}
 \hfil
 \subfloat[\scriptsize V2X-ViT in s-2]{%
  \includegraphics[width=0.45\columnwidth]{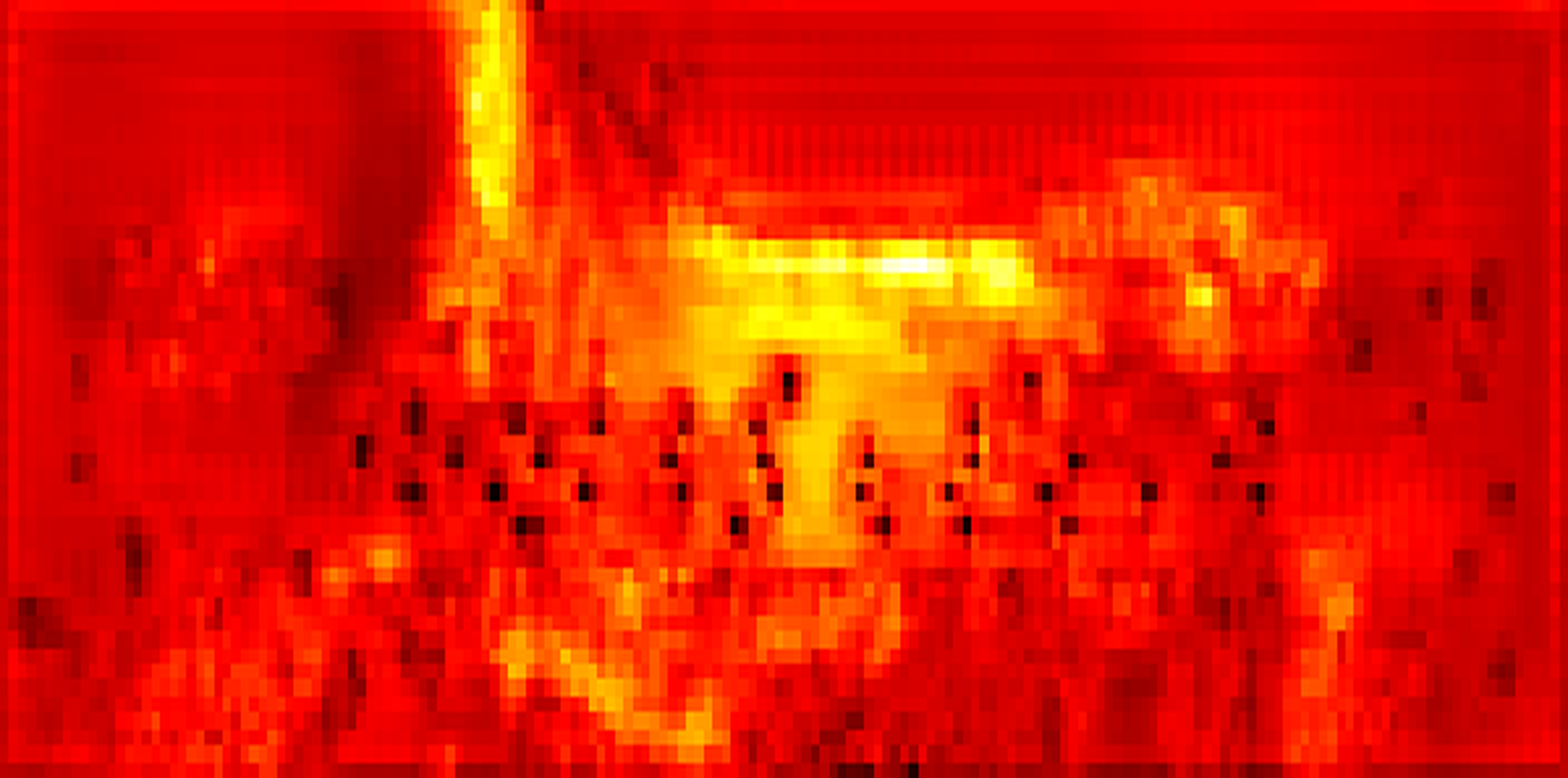}%
}
\\
\subfloat[\scriptsize CoMamba in s-1]{%
  \includegraphics[width=0.45\columnwidth]{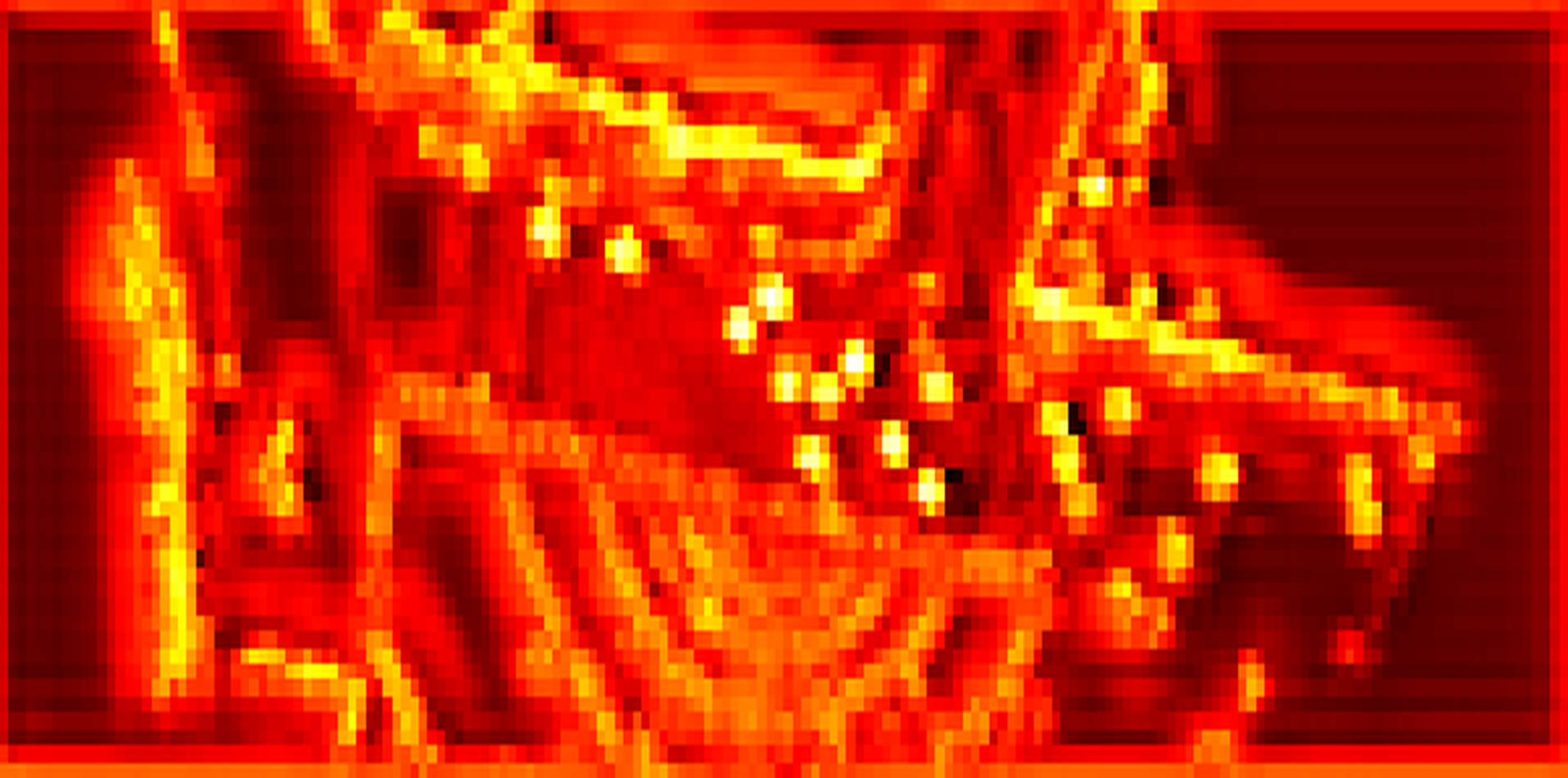}%
}
 \hfil
\subfloat[\scriptsize CoMamba in s-2]{%
  \includegraphics[width=0.45\columnwidth]{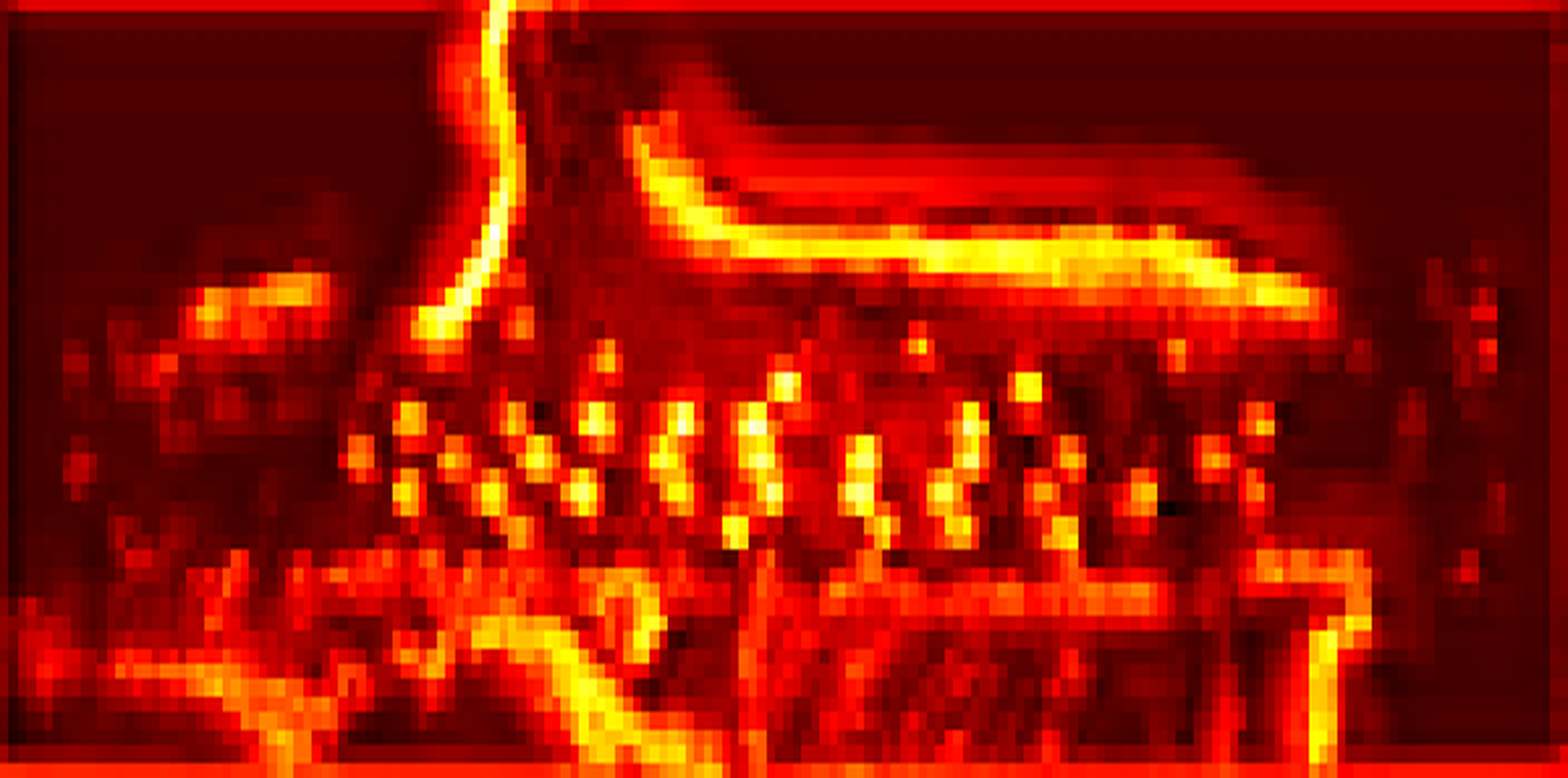}%
}
\captionsetup{font=small}
\caption{\textbf{Visualization of Fused Intermediate Features on the OPV2V Default Testing Set.} We compared the fused intermediate features from our method, CoMamba, against other SOTA methods, V2X-ViT and CoBEVT, using two samples. The first row shows two point cloud samples from our CoMamba, corresponding to the fused intermediate features in the following rows. It is evident that our fused intermediate features are clearer, with more accurate local features corresponding to objects. Additionally, the shape of the scenario modeling is more complete compared to the other methods. }
\label{fig:fea_opv2v}
\vspace{-2.5em}
\end{figure}

\subsection{Quantitative Evaluation}
\label{ssec:quantitative-evaluation}
\noindent \textbf{Results on simulation data.}
As shown in Table~\ref{tab:det_result}, all CP methods significantly surpass the \textit{NO Fusion}, demonstrating the benefits of the V2X perception system on three simulated testing sets.
In the OPV2V Default testing set, our proposed CoMamba outperforms the other seven advanced fusion methods, achieving $91.9\%$/$83.3\%$ for AP@0.5/0.7, highlighted in bold in Table~\ref{tab:det_result}.
In the V2XSet testing set, V2X-ViT~\cite{xu2022v2x} achieves $88.2\%$/$71.2\%$ for AP@0.5/0.7, while our CoMamba attains $88.3\%$/$72.9\%$ for AP@0.5/0.7, surpassing V2X-ViT~\cite{xu2022v2x} with an AP@0.7 improvement of $1.7\%$. These results indicate that our proposed CoMamba can efficiently enhance the interaction between CAVs' features, achieving the best performance in simulated V2X point cloud data.

\noindent \textbf{Results on real-world data.}
The simulated point cloud data does not accurately reflect the challenges encountered in real-world deployment, as shown in Fig.~\ref{fig:v2x_vis}. To address this, we evaluate all fusion methods on the real-world V2V4Real testing set, presented in Table~\ref{tab:det_result}.
%
Our proposed CoMamba outperforms the other seven advanced fusion methods, achieving $63.9\%$/$35.5\%$ for AP@0.5/0.7, which is higher than the second-best fusion method, CoBEVT~\cite{xu2022cobevt}, with an AP@0.5/0.7 improvement of $5.3\%$/$5.8\%$. This indicates that our CoMamba, with its CSS2D and GPM modules, can effectively enhance global interaction capabilities in complex real-world V2X data, resulting in excellent cooperative perception performance.

\noindent \textbf{Results on visual 3D object detection.}
As shown in Table~\ref{tab:vis_result}, our proposed CoMamba model surpasses other advanced fusion methods, delivering superior 3D  perception performance on the camera-only V2X perception system. 

\begin{table}[htb]
\vspace{-4mm}
\centering
\captionsetup{font=small} 
\caption{\textbf{Component ablation study.}
}
\label{tab:ablation}
\resizebox{0.36\textwidth}{!}{%

\begin{tabular}{@{}c|cc|cc@{}}
\toprule
 &
  \multicolumn{2}{c|}{V2XSet~\cite{xu2022v2x}} &
  \multicolumn{2}{c}{V2V4Real~\cite{xu2023v2v4real}} \\ \cmidrule(l){2-5} 
\multirow{-2}{*}{} &
  \multicolumn{1}{c|}{AP@0.5} &
  AP@0.7 &
  \multicolumn{1}{c|}{AP@0.5} &
  AP@0.7 \\ \midrule
  Baseline &
  \multicolumn{1}{c|}{71.4} &
  54.7 &
  \multicolumn{1}{c|}{48.5} &
  \multicolumn{1}{c}{24.1}  \\
\textbf{w/} CSS2D &
  \multicolumn{1}{c|}{86.9} &
  71.5 &
  \multicolumn{1}{c|}{58.1} &
  \multicolumn{1}{c}{33.9}  \\
  \textbf{w/} GPM &
  \multicolumn{1}{c|}{84.4} &
   68.4 &
  \multicolumn{1}{c|}{57.3} &
  \multicolumn{1}{c}{30.5}  \\
CoMamba &
  \multicolumn{1}{c|}{ \textbf{88.3}} &
   \textbf{72.9} &
  \multicolumn{1}{c|}{\textbf{63.9}} &
  \multicolumn{1}{c}{\textbf{35.5}}  \\ \bottomrule
\end{tabular}
}
\vspace{-2.3em}
\end{table}

\noindent \textbf{Efficiency analysis.}
Fig.~\ref{fig:cavnum} illustrates the processing performance comparison with current popular V2X perception methods.
In current V2X datasets, our CoMamba achieves \underline{real-time} perception performance with an inference speed of \textit{26.9 FPS while utilizing only 0.64 GB of GPU memory}. Even when the number of agents increases to 10, CoMamba maintains a solid performance with 7.6 FPS and a GPU memory usage of 7.3 GB.
The linear time complexity of our CoMamba makes it particularly advantageous for real-time 3D perception in real-world, large-scale driving scenarios.

\noindent \textbf{Visualization.}
Fig.~\ref{fig:v2x_vis} presents 3D detection visualization examples from V2X-ViT~\cite{xu2022v2x}, CoBEVT~\cite{xu2022cobevt}, and our CoMamba across three testing sets. It is evident that our proposed CoMamba achieves more accurate 3D detection results in both simulated and real-world point cloud scenarios, demonstrating its superior performance in cooperative perception tasks.
We visually present intermediate features in Fig.~\ref{fig:fea_opv2v} using two point cloud samples. 

\noindent \textbf{Ablation study.}
Table~\ref{tab:ablation} highlights the significance of our proposed CSS2D and GPM within the CoMamba framework on the V2XSet~\cite{xu2022v2x} and V2V4Real~\cite{xu2023v2v4real} testing sets. The baseline is a simple averaging fusion method with a $1\times1$ convolution layer. Integrating CSS2D and GPM into CoMamba resulted in performance improvements of $15.4\%$/$11.4\%$ for AP@0.5/0.7 compared to the Baseline on the V2V4real testing set, underscoring their substantial contribution to the overall performance.


\section{Conclusion} \label{sec:conclusion}

In this paper, we introduce a novel attention-free, state space model-based framework called CoMamba for V2X-based perception.
Our innovative framework incorporates two major components: the Cooperative 2D-Selective-Scan Module (CSS2D) and the Global-wise Pooling Module (GPM), which are responsible for enhancing global interaction efficiently and could be utilized in future large-scale V2X perception scenarios. By leveraging the advantages of SSMs, 
CoMamba enables real-time cooperative perception with an impressive inference speed of 26.9 FPS while utilizing only 0.64 GB of GPU memory footprint.
Furthermore, CoMamba scales remarkably well, achieving linear-complexity costs in GFLOPs, latency, and GPU memory relative to the number of agents, while still maintaining excellent perception performance.
Our extensive experiments on both simulated and real-world V2X datasets demonstrate that CoMamba surpasses other state-of-the-art cooperative perception methods on the 3D point cloud object detection task.
We envision that our work will facilitate novel architectural designs and practical onboard solutions for real-time cooperative autonomy.
 

\bibliographystyle{IEEEtran}
\bibliography{Jinlong}

\end{document}